\documentclass[]{style}

\usepackage{amssymb}
\usepackage{xcolor}
\usepackage{tcolorbox}
\usepackage[utf8]{inputenc}
\usepackage[T1]{fontenc}
\usepackage{lmodern}
\usepackage{wrapfig}
\usepackage{tcolorbox}
\usepackage{array}

\usepackage{array}
\usepackage{makecell}
\usepackage{soul}

\usepackage{graphicx}
\usepackage{tcolorbox}
\usepackage{fontawesome5}
\usepackage{xcolor}
\usepackage{booktabs}
\usepackage{multirow}
\usepackage{algpseudocode}
\usepackage{listings}
\usepackage{enumitem}
\usepackage{caption}
\usepackage{mdframed}
\usepackage{tcolorbox}
\usepackage{algorithm}
\usepackage{algpseudocode}  
\usepackage{amsmath}        
\usepackage{tabularx}
\usepackage{algpseudocode}
\usepackage{amsmath,amsfonts,bm}

\def\eqref#1{equation~\ref{#1}}

\def\1{\bm{1}}

\DeclareMathAlphabet{\mathsfit}{\encodingdefault}{\sfdefault}{m}{sl}
\SetMathAlphabet{\mathsfit}{bold}{\encodingdefault}{\sfdefault}{bx}{n}

\newtcolorbox{takeawaybox}[1]{
    colback=green!5!white,
    colframe=green!50!black,
    fonttitle=\bfseries,
    title=Takeaway
}

\title{\LARGE Geometric Metrics and LLMs:\\
\Large What They Measure and When They Work}

\author[]{Viacheslav Yusupov}
\author[]{Anna Antipina}
\author[]{Ameliia Alaeva}
\author[]{Danil Maksimov}
\author[]{Anna Vasileva}
\author[]{Tatyana Zaitseva}
\author[]{Alina Ermilova}
\author[]{Evgeny Burnaev}
\author[]{Egor Shvetsov }

\abstract{
We present a systematic stress-test of geometric metrics for LLM evaluation. Rank-based geometric properties of internal representations have shown promise as reference-free quality signals, but the conditions under which they are reliable remain unclear. We evaluate eight commonly-used metrics: intrinsic-dimensionality estimators, spectral norms, and related quantities across six tester models (0.5–8B) and eight generators on contrasting tasks, separating genuine geometric signal from text-length effects and from what standard text statistics already capture. Three findings emerge. First, some metrics (notably Schatten Norm and MOM) mainly reflect output length, and their apparent discriminative power collapses once length is controlled. Second, geometric metrics add modest but real information beyond text statistics: combined with them, a classifier reaches 78\% accuracy on 6-way generator identification versus 69\% for text statistics alone. Third, rather than tracking a general notion of text quality, the metrics demonstrate only moderate association between the intrinsic-dimensionality and lexical diversity (RTTR). We give use-case-specific recommendations and identify failure detection as the most promising near-term application.}
\date{\today}

\begin{document}
\maketitle

\section{Introduction}
\label{sec:introduction}

Evaluating large language models (LLMs) remains a fundamental challenge.
The dominant paradigm relies on benchmarking measuring accuracy on
curated annotated datasets which has well-known
limitations~\citep{ni2025survey, hu2024unveiling}: it requires a new
benchmark for each application scenario, prioritizes utility over text
quality, and is increasingly undermined by benchmark contamination in the
training sets of newer models~\citep{ni2025survey}. Alternative metrics
carry their own drawbacks. Reference-based metrics such as BLEU and ROUGE
require human-written references and struggle to capture nuanced semantic
quality~\citep{wang2023automated}. Perplexity, while reference-free, has
recently been shown to be an unreliable quality signal in many practical
settings~\citep{velivckovic2026perplexity}.

Geometric properties of LLM internal
representations have attracted growing interest as candidate
reference-free signals. The motivation is well-founded: in
self-supervised pretraining, rank-based geometric criteria predict
downstream quality well enough to be practically useful. RankMe shows
that the effective rank of learned representations tracks downstream
performance closely enough to select hyperparameters without any
labels~\citep{garrido2023rankme}, and LiDAR refines this into a measure
of linear-probing performance in joint-embedding
architectures~\citep{thilak2024lidar}. Metrics such as Effective Rank,
Intrinsic Dimensionality (ID), and anisotropy measures are likewise
theoretically appealing for text: they are annotation-free, grounded in
the geometry of how models encode language, and have shown promise in
isolated settings~\citep{tulchinskii2023intrinsic, wei2024diff}. \emph{A
recurring and attractive hypothesis is that such metrics can serve as
general-purpose proxies for text quality}.



For a practitioner deciding whether
to build an evaluation pipeline around these metrics, the open questions are
concrete: what do they actually measure on text, under what conditions are
they reliable, and where do they fail?

This paper answers those questions for a practitioner weighing whether to
build on these metrics. Our aim is usefulness rather than novelty: we
cleanly separate results we \emph{reconfirm} from prior and concurrent
work, results where geometric metrics genuinely \emph{help}, and the
settings where they \emph{do not work}. We study eight geometric metrics: MEV, Effective Rank, the intrinsic-dimensionality set
CorrInt/MOM/MLE/MADA, Schatten Norm, and Resultant Length across three
datasets, and four task types, using six \textbf{tester models} - models
that read internal representations and eight \textbf{generator
models} - models that produce the text. For brevity, we use the terms \textbf{tester} and \textbf{generator}
throughout the paper.

To organize the analysis, we apply a short diagnostic: for any representation-based metric, ask whether it
(i)~reads the \emph{text} rather than the measuring model, (ii)~adds
information \emph{beyond} standard text statistics, and (iii)~shows
convergent validity with established external text metrics. 

\textbf{What we reconfirm.} Several established findings replicate
cleanly across our testers and tasks, and we treat them as background
rather than claims. (1)  Intrinsic-Dimensionality (ID) separates human from AI-generated
text~\citep{tulchinskii2023intrinsic}, (2) the ID set of metrics 
follows the mid-depth ``hunchback'' profile across tester models
we study~\citep{viswanathan2025geometry}, (3) and ID's clearest interpretable
association is with lexical diversity (type--token ratio), independently and
concurrently reported by \citet{pedashenko2025unveiling}. 


\textbf{What works.} Geometric metrics add
real but modest discriminative signal. Combined with standard text
statistics they improve 6-class model-generator identification from $0.69$ to
$0.78$ accuracy, and in human-vs. AI detection a strong tester (Qwen2.5-7B)
reaches PR-AUC $=0.95$ versus a $0.89$ text statistics
baseline--though sharply dependent on a capable tester model.

\begin{takeawaybox}

When labeled data and standard text
statistics are available, those statistics are the stronger baseline, and
geometric metrics are a small complement rather than a replacement.
\end{takeawaybox}

\textbf{What does not work (and why).} We report the negative results as
a contribution in their own right, since knowing where an approach fails
saves real effort. We find that geometric metrics are largely driven by
sequence length rather than by any property of the writing: Schatten Norm
and MOM are the most affected (Spearman $\rho > 0.7$ with length on
Generated text), while Effective Rank and MEV are the most
length-independent. Before drawing any conclusion we therefore residualize
each metric against length to separate genuine geometric signal from a
length artifact.

This control changes the picture substantially. Most importantly,
geometric metrics are \emph{not} general-purpose quality proxies: once
length is removed, no metric aligns with external text metrics (PPL,
BLEURT, RTTR, semantic diversity, compression rate). The only signal that
survives is MOM and CorrInt tracking RTTR ($\rho \approx 0.54$ on Generated
text), so these metrics are best understood as proxies for lexical
diversity, not as quality measures. The length confound also explains the
strongest apparent results,

Beyond length, three further factors appear to degrade these metrics.
First, \emph{the measuring model must be strong enough}: weak testers
($\leq 2$B) give noisy, inconsistent rankings. Second,
\emph{detectability is language-dependent}: the human-vs-AI separation
that is clear in English is substantially weaker in a lower-resource
language (Russian). We note this cross-lingual pattern is based on visual
inspection of the layer-wise profiles
(Figure~\ref{fig:crosslingual_layers}) rather than on a quantitative
detection score, and report it as a qualitative observation. Third,
\emph{sensitivity to quality degradation is coarse}: geometric metrics
flag quantization damage only when it is severe---CorrInt drops for 2-bit
generation but does not separate 4-bit from full precision
(\S\ref{sec:quant_main})---so they miss the moderate precision reductions
most common in practice.

Throughout our experiments, we aggregate each metric
by averaging values over model layers, this was a design choice we adopted at the
outset rather than one we validated, and our results should be read with
that in mind. The layer-wise profiles
(Figures~\ref{fig:layer_profiles_1}--\ref{fig:layer_profiles_2}) show that
the signal is distinctly non-uniform across depth, so averaging is
unlikely to be optimal. We therefore expect a more intelligent
depth-aware aggregation (e.g.\ a mid-layer band) to recover signal our
layer-mean dilutes, and we flag this as a promising direction rather than
a settled result.


\section{Background and Related Work}
\label{sec:related-work}
\textbf{Geometric properties} such as Intrinsic Dimension (ID) applied to text embeddings exhibit varying  values for different languages, and ID is consistently lower for AI-generated text compared to human-written content~\citep{tulchinskii2023intrinsic}. By applying sparse autoencoders to LLMs residual streams, \citep{kuznetsov2025feature} has demonstrated that LLMs produce text with distinctive stylistic features and that text generated by different model classes can be distinguished from one another. \citet{viswanathan2025geometry} showed that ID is higher in middle layers of LLMs compared to other layers, and that ID for randomized text is significantly higher indicating models struggle to compress it. Intrinsic dimensionality across different layers has been considered as a measure of generalization capability in both convolutional networks \citep{ansuini2019intrinsic} and LLMs \citep{roy2007effective}. \citet{godey2024anisotropy} demonstrated that anisotropy is an inherent property of transformer-based models and can itself serve as a model characteristic, potentially useful for detecting hallucinations \citep{yin2024characterizing}. The works \cite{wei2024diff} and \cite{deng2025unify} also propose using effective rank for model evaluation. A complementary line of work finds that geometric signals are most reliable when paired with a \emph{specific, operationalized} target rather than a diffuse notion of quality: \citet{nguyen2025distance} use the radial dispersion of sampled generations in embedding space as a state-of-the-art signal for uncertainty and hallucination detection, and \citet{schiekiera2026associations} show that hidden-state semantic geometry is recoverable from behavioral similarity judgments. These results motivate our framing: we ask not whether geometric metrics track ``quality'' in general, but against which concrete external measures they show convergent validity.

\textbf{The evaluation of text quality} typically employs various metrics. GPT-2 Perplexity is commonly used to assess the fluency and naturalness of generated text by measuring how well a pre-trained model predicts the text sequence \citep{chang2024scaling}. Traditional metrics like BLEU and ROUGE measure lexical similarity to human-written references, though they struggle to capture nuanced semantic aspects \citep{wang2023automated}. The perplexity metric is also have problems with model evaluations \cite{velivckovic2026perplexity}. At the same time, BLEURT~\citep{yan2023bleurt} demonstrates higher correlation with human judgments of overall text quality capturing semantic adequacy and entailment, along with aspects of fluency and grammar. 
Other approaches include using compression ratios with algorithms like zip to estimate text diversity \citep{chang2024scaling} and examining statistical differences in linguistic features between human and LLM-generated texts, including average subtree height, dependency tree height, and sentence length \citep{yu2024towards}.

\section{Geometric Metrics: Definitions and Motivation}
\label{sec:metrics}

Transformer-based language models build  abstract 
representations of input text as it passes through successive layers. 
The geometry of these internal representations reflects properties of 
the text itself rather than of any particular downstream task. This 
makes geometric metrics a natural candidate for reference-free, 
annotation-free text evaluation.

These representations also evolve in a characteristic way with depth. A
growing body of work reports a universal ``expansion--compression'' or
\emph{hunchback} profile of intrinsic dimensionality across layers: it is
low in early layers, peaks in the middle of the network where abstraction
is greatest, and compresses again toward the output~\citep{viswanathan2025geometry,
pedashenko2025unveiling}. We observe exactly this profile for the
intrinsic-dimensionality cluster across every tester model we study
(Figures~\ref{fig:layer_profiles_1} \ref{fig:layer_profiles_2}), and as we show
in \S\ref{sec:practical} the separation between generators is
concentrated at this mid-depth peak. This is the geometric region we
expect to be most informative for evaluation, and it motivates reading
these metrics with attention to depth rather than as a single
layer-averaged scalar.

\subsection{Geometric Metrics and Why We Use Them}
\label{sec:methodology}
We  select metrics that have prior theoretical or empirical motivation for 
capturing systematic differences between human-written and 
generated text~\citep{tulchinskii2023intrinsic, razzhigaev2024shape, 
godey2024anisotropy}, and that are applicable to any 
transformer-based model without modification.

The selected metrics cover complementary geometric properties: 
directional concentration of token embeddings (MEV, Resultant 
Length), spectral complexity of the representation space (Effective 
Rank, Schatten Norm), intrinsic manifold structure (Intrinsic 
Dimensionality), and distributional alignment with human text 
(MAUVE). Metrics and their brief description is summarized in Table~\ref{tab:metrics_summary}, full description of the metrics is provided in Appendix~\ref{app:metrics}. 

\begin{table}[htbp]
\centering
\caption{Summary of evaluation metrics used in this work. Direction 
indicates whether higher ($\uparrow$) or lower ($\downarrow$) values 
are generally preferable for natural, diverse text.}
\label{tab:metrics_summary}
\resizebox{\textwidth}{!}{%
\begin{tabular}{lcl}
\toprule
\hline
\textbf{Metric} & \textbf{Direction} & \textbf{Brief Description} \\
\hline
Schatten Norm & --- & Global spectral energy, measures magnitude/stability. \\
MEV & $\downarrow$ & Fraction of variance in top singular values. \\
Resultant Length ($R$) & $\downarrow$ & Bigger value means all vectors point into the same direction. \\
Effective Rank (ERank) & $\uparrow$ & Entropy of normalized singular values. \\
MAUVE & $\uparrow$ & Distributional alignment to human text in feature space. \\
Intrinsic Dimension (ID) & $\downarrow^*$ & Estimated manifold dimension --- $^*$lower = more predictable. \\
\hline
\end{tabular}%
}
\end{table}

\section{Experimental Setup}
\label{sec:setup}

A key question we investigate is whether geometric metrics, despite their 
different geometric motivations, produce consistent signals across 
tester models and tasks  and what they actually measure when they 
do.

\subsection{Tester/Generator Framework}
\label{framework}
We evaluate text using two disjoint sets of models: \textbf{tester 
models} $\mathcal{T}$, which serve as measurement instruments, and 
\textbf{generator models} $\mathcal{G}$, which produce the text being 
evaluated. For each generator $g \in \mathcal{G}$, we pass its outputs 
through each tester $t \in \mathcal{T}$ and extract hidden states 
$X^{(l)}_g \in \mathbb{R}^{n \times d}$ at each layer $l$, where $n$ 
is the sequence length and $d$ the hidden dimension. States are taken 
from MLP blocks after activation functions and before residual 
connections. For a metric $R: \mathbb{R}^{n \times d} \rightarrow 
\mathbb{R}$, we compute a scalar summary by averaging over all $L$ 
layers: $s^{R}(X_g) = \frac{1}{L}\sum_{l=1}^{L} R(X_g^{(l)}) \in 
    \mathbb{R}$.
This scalar serves as the primary signal for comparing generators in 
$\mathcal{G}$.

We use six tester models ($\mathcal{T}$): Gemma-1-7b, 
Gemma-2b-it, LLaMA-3.1-8B-it, Qwen-2.5-7b-it, Qwen-2-0.5B, and 
LLaDA-8B; and eight generator models ($\mathcal{G}$): Gemma-2b-it, 
Qwen-2.5-7b-it, LLaMA-3.1-8B-it, DeepSeek-R1, Mistral-7b-it, 
Phi-3-medium-4k-it, Phi-3-mini-128k-it, and Starling-lm-7b-beta. 
For the quantization experiments, we additionally include LLaMA and 
Mistral models under three precision levels: full precision (fp32), 
4-bit quantization (AWQ), and 2-bit quantization (AQLM), yielding 
five quantized variants in total. Instruction prompts for all of these settings are provided in Appendix~\ref{app:prompts}.

\subsection{Datasets, Tasks, and Generation Modes}
\label{seq:datasets}

We evaluate geometric metrics across three datasets and four 
generation modes, chosen to cover a range of conditions under which the metrics might behave differently. Firstly, we evaluate $1000$ original and \textbf{rewritten film reviews} in English (IMDB dataset~\citep{imdb2011}), Russian (Kinopoisk 
reviews~\citep{klemin_kinopoisk_2024}) and German (German sentiment 
corpus~\citep{guhr-EtAl:2020:LREC}) languages. 

Secondly, we utilize the abstracts from arXiv dataset~\citep{clement2019arxiv}. Each abstract is processed in 
two distinct generation modes: \textbf{(1) Simplification:} generator models rewrite the  abstract for a general audience, preserving the core message 
    while reducing technical complexity.
\textbf{(2) Title-based generation:} generator models produce 
    a plausible academic abstract given only the paper title, 
    requiring open-ended generation with no source text to 
    paraphrase. 

To assess whether geometric metrics can detect compression-induced 
quality degradation, we include outputs from LLaMA and Mistral 
models under different quantization 
schemes~\citep{kharinaev2025investigating}, ranging from full 
precision (fp32) to 4-bit (AWQ) and 2-bit (AQLM) quantization.

Instruction prompts for all of these settings are provided in Appendix~\ref{app:prompts}.

\subsection{Baseline Text Quality Metrics}
\label{sec:text-metrics}

We compare geometric metrics against a broad suite of established 
text quality metrics covering complementary aspects of text quality. 
For reference-based evaluation we use ROUGE-L and BLEURT,  
which compare generated text against a reference. For reference-free 
evaluation we use GPT-2 perplexity (PPL), compression rate (CR) via 
ZIP, Root Type-Token Ratio (RTTR), semantic diversity (SemDiv), and 
Flesch Reading Ease (FRE). Together these metrics capture fluency, 
semantic faithfulness, lexical diversity, and readability. Full 
definitions and preprocessing details are provided in 
Appendix~\ref{appendix:text-metrics} and the results of text metric evaluation are provided in Table \ref{tab:metrics_comparison}.


\section{Results and Experiments}
\label{sec:results}
\vspace{-0.5em}

In this section, we provide the results of our experiments for different tasks, models and metrics.

\paragraph{Are geometric metrics practically useful, and when?}
\vspace{-0.5em}
\label{sec:practical}
This section evaluates the metrics against three concrete
practical criteria: \textbf{(T1)} does the
metric detect properties of the text or of the tester model, \textbf{(T2)} does adding a geometric metric to a model
trained on length, perplexity, and lexical-density features improves generator model
classification, and \textbf{(T3)} does the metric show convergent validity
with a panel of established external text metrics? 


\textbf{Length control.} Generators produce systematically different output lengths as depicted in Figure~\ref{fig:p1-lengths} in Appendix \ref{app:text_length}. Overall, the average human-written text length equals $135$ tokens, whilst for different LLMs the length is vary significantly from $135$ to up to $200$ tokens in simplification scenario and up to $250$ tokens in Generation one. To isolate length-independent signals, we analyze both raw metrics and length-residualized~\footnote{We
fit a linear regression on output length within each task and subtract the
length-dependent component, preserving column means so the structure of
the analysis is preserved} values.  A metric whose value  collapses after residualization is one whose apparent signal is largely length.


\textbf{Does a metric read the text or the instrument? [BOTH]}
A metric is only a useful evaluation signal if its value is driven by
the text being measured rather than by the model used to measure it. We
make this question quantitative by asking how a metric's total variance
splits between the axis we care about (the text) and the axes we do not
(the generator that wrote it, the probe model that reads it). We use a
fixed-effects variance decomposition as the natural instrument for this
partition.

For each task and metric $R$, we build a balanced paragraph~$\times$~generator
table $X_{ij} = R(\text{paragraph}_i,\, \text{generator}_j)$, taking each
cell as the average of $R$ over our three testers (Gemma 7B, Qwen2.5 7B,
LLaMA-3.1 8B) so that the entry reflects signal common to capable probes
rather than any single one. We then split the total variance of this table
into a generator term, a paragraph term, and a residual:
\begin{equation}
\underbrace{\sum_{i,j}(X_{ij}-\bar X)^2}_{SS_{\text{total}}}
= \underbrace{N \sum_j (\bar X_{\cdot j}-\bar X)^2}_{SS_{\text{generator}}}
+ \underbrace{J \sum_i (\bar X_{i\cdot}-\bar X)^2}_{SS_{\text{paragraph}}}
+ SS_{\text{residual}},
\label{eq:decomp}
\end{equation}
which are mutually orthogonal in a balanced design. A large generator share
means the metric separates models, a large paragraph share means it tracks
the text. We clip each metric to its $1\%$--$99\%$ range within each task to
limit the influence of outliers.

\begin{figure}[t]
  \centering
  \includegraphics[width=\linewidth]{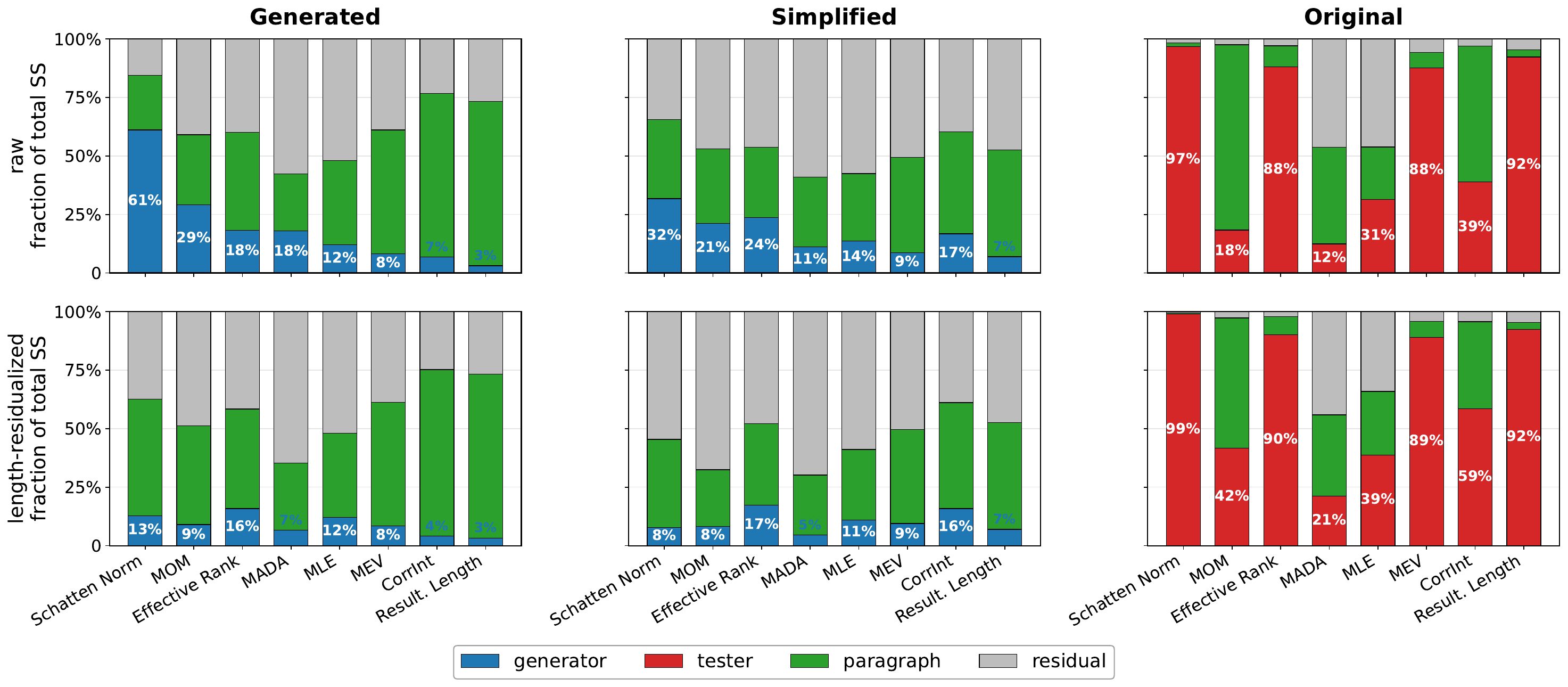}
  \caption{Variance decomposition of geometric metrics, before
  (\emph{top row}) and after (\emph{bottom row}) length residualization.
  \textbf{Left, middle} columns: AI text, blue bar = generator share
  (larger = stronger generator-discriminator).
  \textbf{Right} column: human text, red bar = tester share (smaller =
  better; we want paragraph variance, not probe-model idiosyncrasy).}
  \label{fig:p2-decomp}
\end{figure}

Human-written text has no generator axis, so we apply the same split with
the testers as the second axis: $X_{ik} = R(\text{paragraph}_i,\,
\text{tester}_k)$ over the $K=3$ testers. This keeps all three panels of
Figure~\ref{fig:p2-decomp} comparable, with the nuisance axis being the
generator on AI text and the tester on human text.

 Variance decomposition (Figure~\ref{fig:p2-decomp}) reveals a nuanced relationship between geometric signals and sequence length. On \textsc{generated} text, metrics such as Schatten Norm, MOM, and MADA lose the majority of their generator-specific signal after length control, indicating that their discriminative power is predominantly length-driven. However, \textbf{Effective Rank} ($18\%\to16\%$) and \textbf{MEV} ($8\%\to8\%$) remain remarkably stable, suggesting they capture other than length properties. This pattern holds for \textsc{simplified} text, where Effective Rank emerges as the most robust length-independent discriminator ($17\%$ variance share). Conversely, results on human text show that variance is overwhelmingly dominated by the \textbf{tester model} ($88$--$97\%$), with residualization causing minimal shifts ($\leq3$ pp).  
 
 \begin{takeawaybox}
 
 This high sensitivity to the choice of probe model suggests that geometric metrics reflect an interaction between the text and the instrument rather than an absolute property of the text alone. To practically employ these metrics, one must explicitly account for both sequence length and the bias introduced by the choice of tester model.
\end{takeawaybox}


\begin{figure}[htb!]
  \centering
  \includegraphics[width=0.8\linewidth]{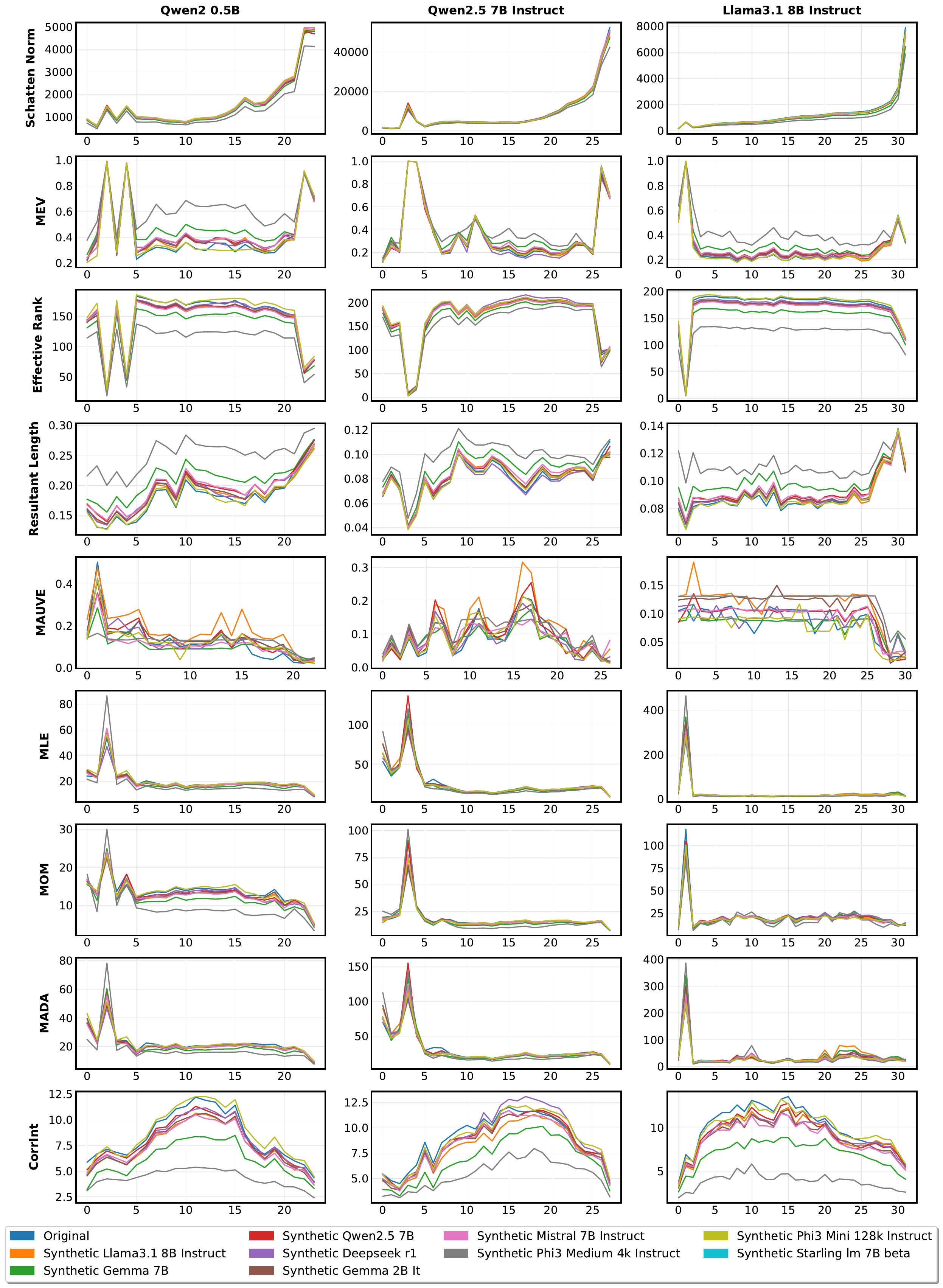}
  \caption{Layer-wise geometric metric profiles for all generators,
  read by three testers (Qwen2-0.5B, Qwen2.5-7B, Llama-3.1-8B). Each
  curve is one generator; \emph{Original} (blue) is human text. Note the
  pronounced mid-depth peak of the intrinsic-dimensionality cluster
  (CorrInt, MLE, MOM, MADA) where generators separate most, and the
  terminal-layer growth of Schatten Norm and edge spikes of MEV that
  dominate any layer-mean.}
  \label{fig:layer_profiles_1}
\end{figure}
\begin{figure}[htb!]
  \centering
  \includegraphics[width=0.8\linewidth]{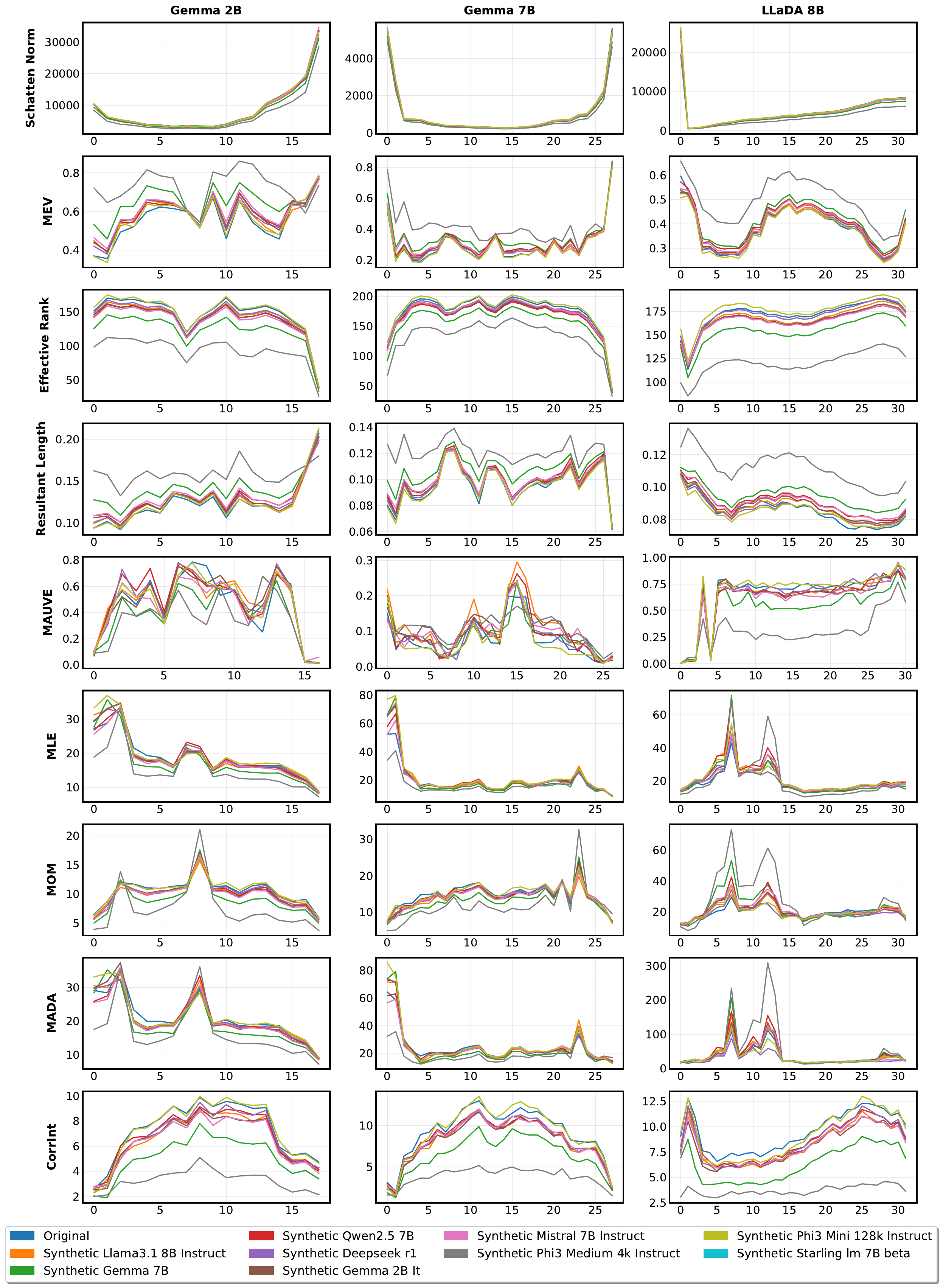}
  \caption{Layer-wise geometric metric profiles read by three further
  testers (Gemma-2B, Gemma-7B, LLaDA-8B), including the diffusion model
  LLaDA. The same depth structure recurs across architectures: a
  mid-depth intrinsic-dimensionality peak that carries the
  generator-discriminative signal, and boundary-dominated Schatten Norm
  and MEV.}
  \label{fig:layer_profiles_2}
\end{figure}
\begin{figure}[ht!]
    \centering
    \includegraphics[width=0.8\linewidth]{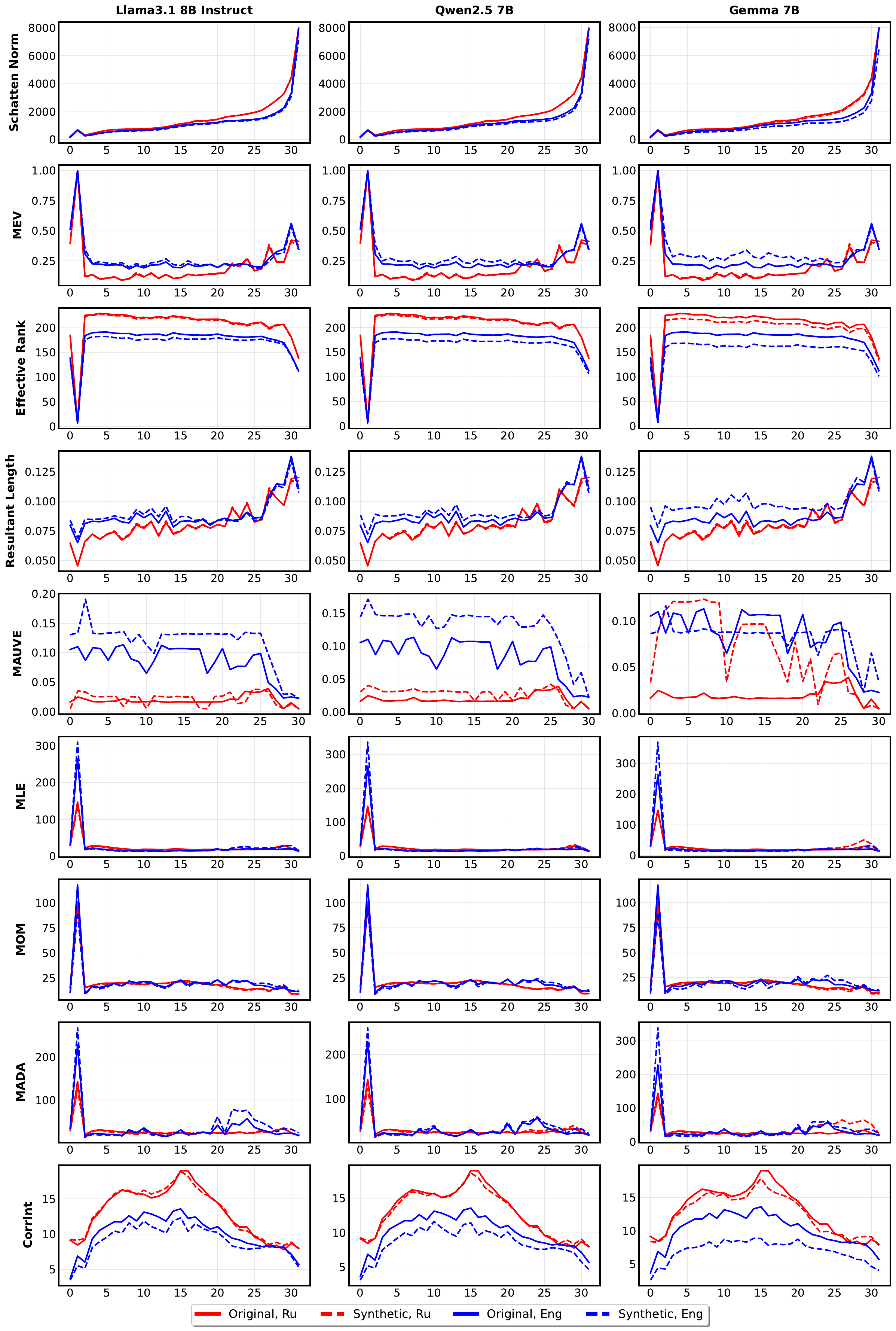}
    \caption{Layer-wise geometric profiles for Russian (red) and English
    (blue), original (solid) vs.\ synthetic (dashed), read by three strong
    testers (Llama-3.1-8B, Qwen2.5-7B, Gemma-7B). Three patterns recur
    across testers: (i) Effective Rank and CorrInt separate \emph{languages}
    more than they separate original from synthetic; (ii) the
    original-vs-synthetic MAUVE gap is large in Russian but collapses in
    English; (iii) the cross-lingual signal is carried by Effective Rank
    and the intrinsic-dimensionality cluster.}
    \label{fig:crosslingual_layers}
\end{figure}

\textbf{Layer-wise profiles.}
The layer-wise profiles
(Figures~\ref{fig:layer_profiles_1}--\ref{fig:layer_profiles_2}) are
strikingly consistent across models: each metric follows the same
characteristic depth pattern in every tester, in line with prior reports
of a universal mid-depth peak in intrinsic
dimensionality~\citep{viswanathan2025geometry, pedashenko2025unveiling}.
This regularity suggests that, regardless of the specific model, particular
layers take on particular functions a picture often described in terms of
an information bottleneck and progressive signal compression, with
representations expanding toward the middle of the network and compressing
again toward the output.

\textbf{Layer averaging.} This depth structure also has a practical
consequence for our layer-averaged scalar. The intrinsic-dimensionality
metrics concentrate their discriminative signal at the mid-depth peak,
while Schatten Norm and MEV are dominated by large boundary-layer values,
so averaging over all layers blends these very different regions,
it attenuates useful signal and partly produces the length confound
(terminal-layer growth scales with sequence length). A depth-aware summary
(e.g.\ a mid-layer band) is a promising, low-cost refinement we leave to
future work.

\textbf{Do geometric metrics add value beyond text statistics (yes)?} A practitioner already has access to standard text statistics: perplexity (PPL), lexical density (RTTR), readability, semantic diversity,
compression rate, and word count. The relevant question is not  does
this geometric metric carry signal but does it carry signal that the
text statistics do not already provide. We test this with a 6-class
generator-identification task: predict which generator (out of
\textsc{deepseek}, \textsc{gemma}, \textsc{llama}, \textsc{qwen},
\textsc{phi-medium}, \textsc{starling}) produced a given output, using a
random-forest classifier trained on either text statistics alone or text
statistics augmented with one or all geometric metrics. We report
5-fold cross-validation accuracy.
\begin{figure}[t]
  \centering
  \includegraphics[width=0.95\linewidth]{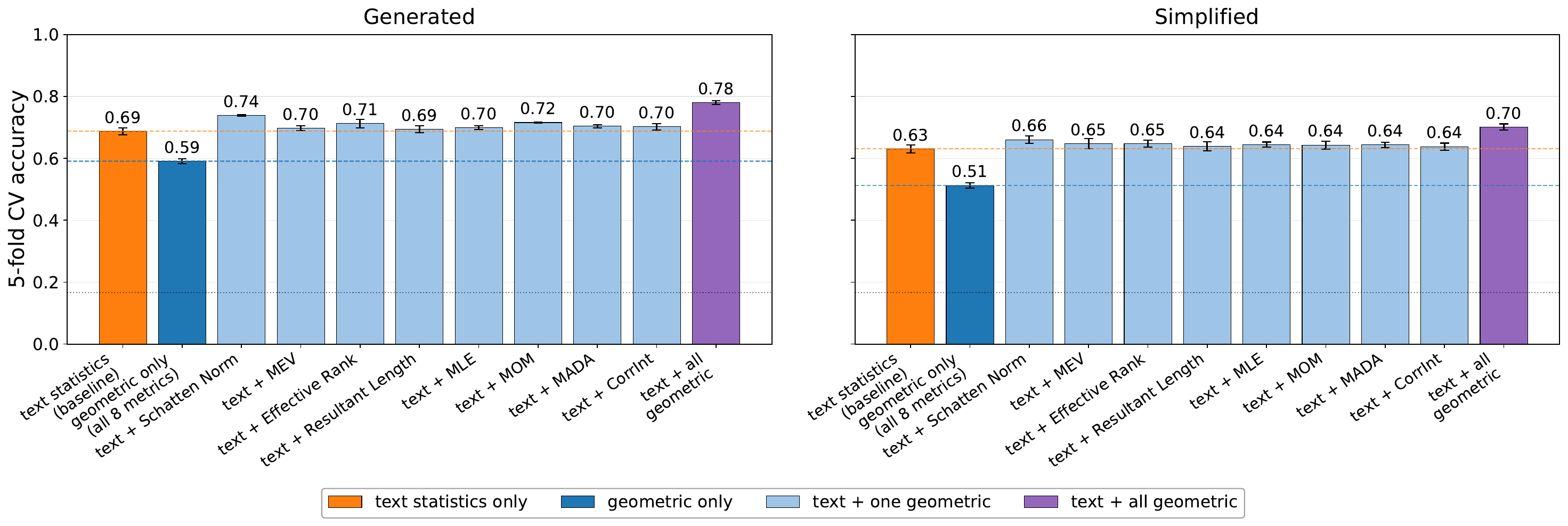}
  \caption{6-class generator-identification accuracy. Orange:
  text-statistics baseline (PPL, RTTR, Readability, Semantic Diversity,
  Compression Rate, word count).  Dashed line represents a  baseline accuracy.}
  \label{fig:p3-added}
\end{figure}

Figure~\ref{fig:p3-added} shows accuracy across feature sets. The text-statistics baseline reaches $0.687$ accuracy on \textsc{generated}
and $0.631$ on \textsc{simplified} already well above chance ($0.167$).
Adding all eight geometric metrics together lifts accuracy to $0.780$ and
$0.701$, respectively. No
single metric adds more than $5$ pp on its own. The strongest individual
contribution comes from Schatten Norm on \textsc{generated} (+$5.2$ pp),
followed by MOM (+$2.9$ pp) on \textsc{simplified} the strongest is again
Schatten Norm (+$2.9$ pp).

 \begin{takeawaybox}

Geometric metrics carry a small amount of generator-discriminative information not captured by standard text statistics. Combining the two feature sets yields the strongest classifier, even though the precise nature of the signal they carry remains an open question.
\end{takeawaybox}

\textbf{Do geometric metrics or text statistics allow us to differentiate between generated and human text?}

Can our research be useful in practice? We test this with binary classification task, predict whether a text is human or generated, using a random-forest classifier trained on either text statistics alone or geometric metrics alone. We report PR-AUC for models trained using all metrics and using each one separately to assess the contribution of each metric to the classification. Dataset: 1000 human texts, 12000 generated texts. 
\begin{table}[htbp]
\centering
\caption{Comparison of geometric metrics by PR-AUC for Human vs Generated Text Classification}
\label{tab:metrics_comparison_classification}
\resizebox{\textwidth}{!}{%
\begin{tabular}{lccccccccc}
\toprule
\textbf{Model} & 
\textbf{all} & 
\textbf{schatten} &
\textbf{anis} & 
\textbf{mle} & 
\textbf{effective} & 
\textbf{corrint} &
\textbf{mada} &
\textbf{mom} &
\textbf{resultant} \\
\midrule
Qwen2.5-7b-it,    & 0.9518 & 0.3015 & 0.1580 & 0.1086 & 0.1444 & 0.1987 & 0.0957 & 0.1741 & 0.0819 \\
LLaDA-8B          & 0.9143 & 0.1510 & 0.1070 & 0.1350 & 0.1539 & 0.1131 & 0.1300 & 0.1668 & 0.1399 \\
Qwen-2-0.5B     & 0.8128 & 0.1402 & 0.1768 & 0.2647 & 0.1234 & 0.0923 & 0.2178 & 0.1222 & 0.1452 \\
Gemma-2b-it     & 0.7929 & 0.2141 & 0.1492 & 0.1860 & 0.1605 & 0.1079 & 0.1786 & 0.1317 & 0.1860 \\
LLaMA-3.1-8B-it & 0.7015 & 0.2282 & 0.3384 & 0.0755 & 0.2734 & 0.1710 & 0.0790 & 0.1400 & 0.0793 \\
Gemma-1-7b      & 0.6806 & 0.1828 & 0.1128 & 0.1559 & 0.1672 & 0.2735 & 0.1936 & 0.1613 & 0.0785 \\

\midrule
\textbf{Mean}   & \textbf{---} & \textbf{0.1906} & \textbf{0.1783} & \textbf{0.1674} & \textbf{0.1643} & \textbf{0.1568} & \textbf{0.1556} & \textbf{0.1478} & \textbf{0.1117} \\
\bottomrule
\end{tabular}%
}
\end{table}

As we can see from the results in Table~\ref{tab:metrics_comparison_classification} and Table~\ref{tab:statistics_comparison_classification}, the geometric metrics captured by Qwen2.5-7b-it and LLaDA-8B perform better than others, including text metrics. The overall result is that both geometric metrics and text statistics have potential for detecting generated text. PPL, RTTR and Schatten norm are the most significant metrics.

\begin{table}[htbp]
\centering
\caption{Comparison of text statistics by PR-AUC for Human vs Generated Text Classification}
\label{tab:statistics_comparison_classification}
\small{%
\begin{tabularx}{\textwidth}{@{\extracolsep{\fill}}ccccc@{\extracolsep{\fill}}}
\toprule 
{\textbf{all}}  & 
{\textbf{PPL}}& 
{\textbf{RTTR}} & 
{\textbf{Semantic diversity}} &
{\textbf{Readability}} \\
\bottomrule
 0.8903 & 0.5778 & 0.5773 & 0.1760 & 0.1075\\
\hline
\end{tabularx}%
}
\end{table}

\textbf{Alignment with external text metrics (weak).} Rather than asking whether
geometric metrics track ``quality'' in the abstract, we ask how well they
align with five established external metrics, each measuring a specific
property: PPL, BLEURT, RTTR, Semantic Diversity, and Compression Rate. We
compute partial Spearman correlations, length-residualized to remove
spurious length-mediated effects, and report reference-based metrics only
where reference text exists.

\begin{figure}[ht]
  \centering
  \includegraphics[width=0.95\linewidth]{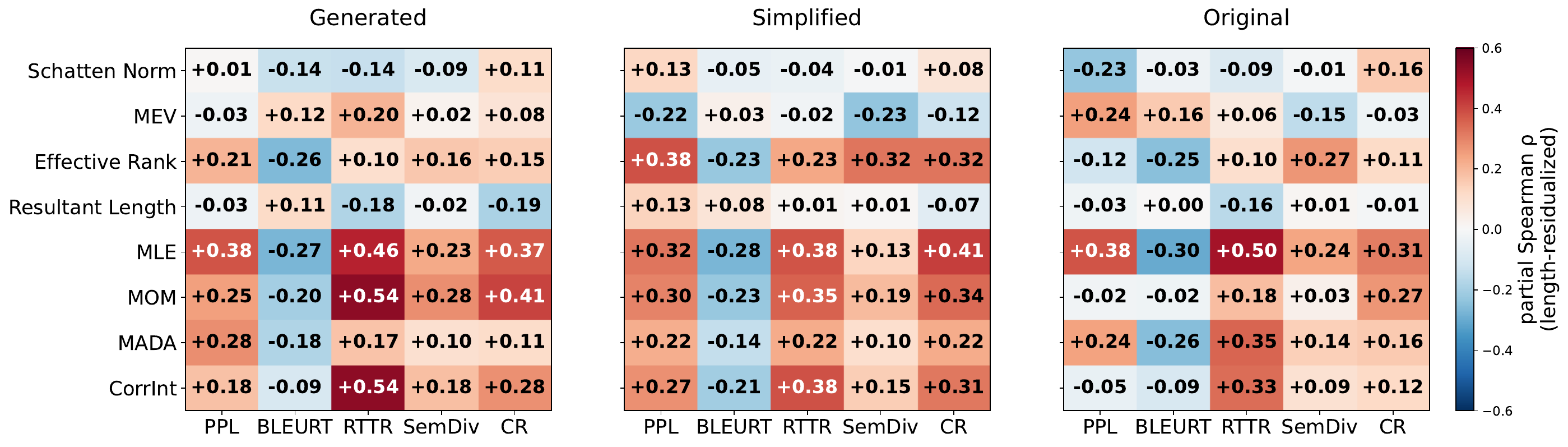}
  \caption{Length-residualized partial Spearman correlation between
  geometric metrics and standard text-quality metrics, by task. Cell
  values are $\rho$ after partialling out word count.}
  \label{fig:p4-conv}
\end{figure}

As shown in Figure~\ref{fig:p4-conv}, most correlations are weak
($|\rho| < 0.3$): geometric metrics largely capture information distinct
from this panel, and no metric aligns consistently across it. Three signals
stand out. First, on \textsc{generated} text, MOM and CorrInt correlate
moderately with lexical diversity (RTTR, $\rho \approx 0.54$), weakening on
other tasks; this is the clearest alignment we observe, and it matches
concurrent work reporting that intrinsic dimensionality tracks type--token
ratio~\citep{pedashenko2025unveiling}. Second, MLE is the most consistent
across metrics and tasks, though still modest. Third, Schatten Norm, a
good generator-discriminator, shows near-zero correlation with
every metric once length is controlled ($|\rho| \leq 0.23$), so its
discriminative power reflects none of these properties. Overall, geometric
metrics behave as selective proxies for lexical diversity, not as quality
measures.

\vspace{-1em}

\begin{figure}[ht!]
    \centering
    \includegraphics[width=0.95
    \linewidth]{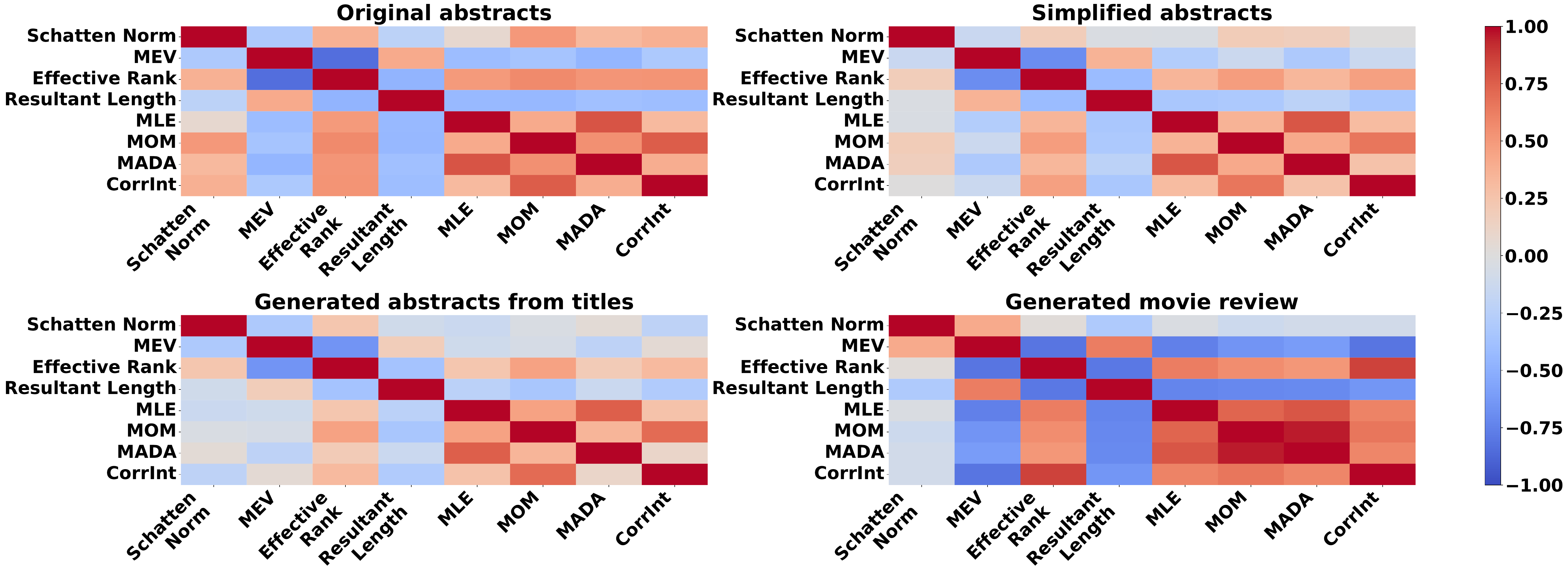}
    \caption{Geometric Metrics correlation with each other for four language tasks (Original abstracts, Simplified abstracts, Generated from titles and Rewritten movie reviews) for all Tester models $\mathcal{T}$.}
    \label{fig:geom_metrics_corr}
\end{figure}
\begin{figure}[ht!]
    \centering
    \includegraphics[width=0.85
    \linewidth]{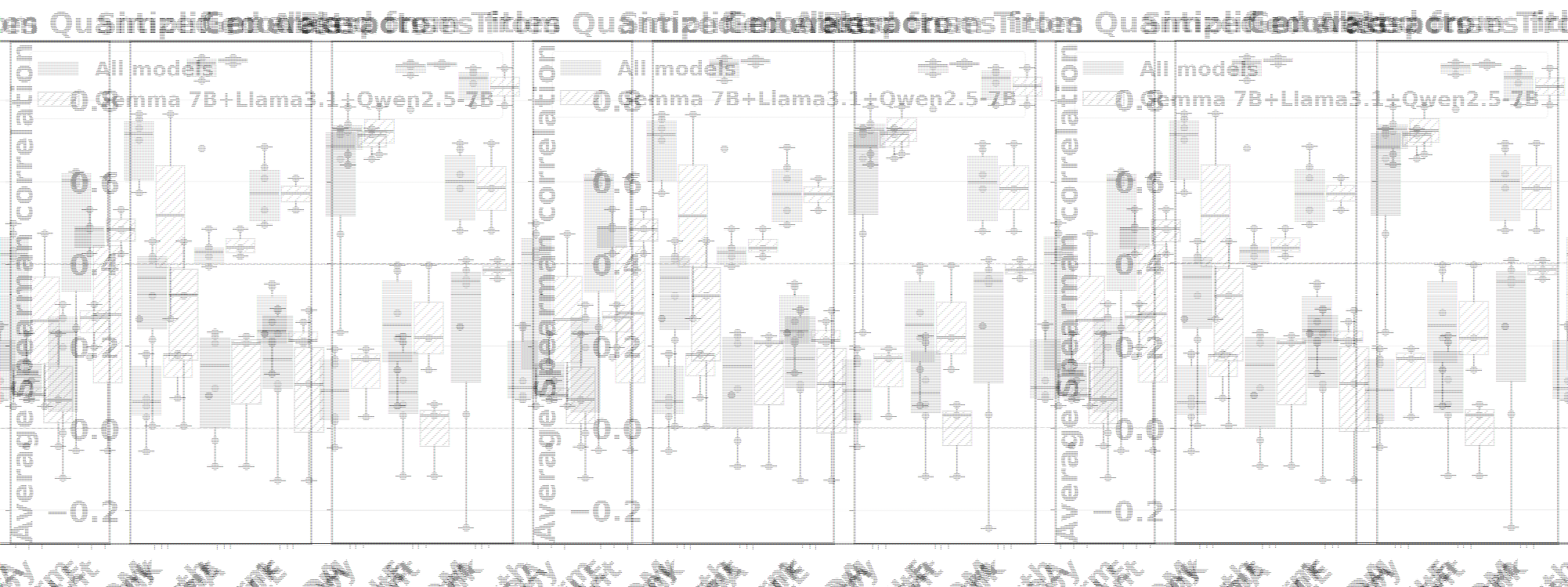}
   \caption{How much do different tester models agree on the ranking of generators? For each geometric metric, we rank the generators using each tester, then measure how similar those rankings are across testers (Spearman correlation). Higher values mean testers agree, lower values mean they disagree on average. Solid boxes use all testers, shaded boxes use only the 7--8B testers (Gemma 7B, Llama3.1, Qwen2.5-7B).}
    \label{fig:correclation_for_geom}
\end{figure}

\begin{figure}[hbt!]
    \centering
    \includegraphics[width=0.9\linewidth]{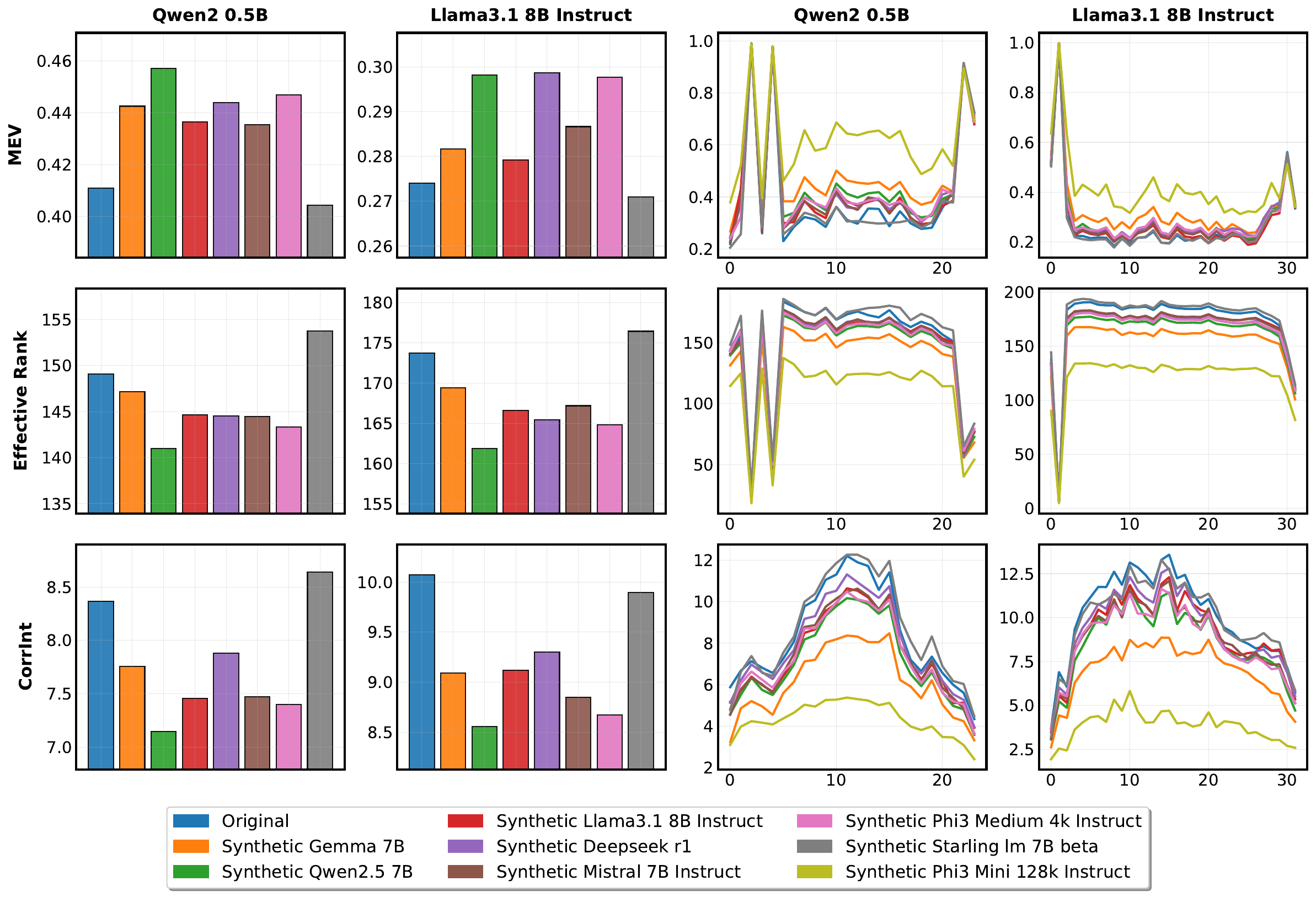}
    \caption{The average across layers (left two columns) and layer-wise (right two columns)  metrics: MEV, Effective Rank and CorrInt for with text generated by various models $\mathcal{G}$ and tester models $\mathcal{T}$: Qwen2 0.5B and Llama3.1 8B Instruct. Original means human written text, while all generated text simply represents rewritten via LLMs original text, which preserves semantic meaning.}
    \label{fig:layer_tot}
\end{figure}

\textbf{When Do Geometric Metrics Agree? The Role of Tester Capacity and Task.}
\label{model_capacity}
We first compute correlations between the geometric metrics across
the four tasks (Figure~\ref{fig:geom_metrics_corr}). The
intrinsic-dimensionality metrics (MLE, MOM, MADA, CorrInt) and Effective
Rank form a tight, highly-correlated cluster on every task, indicating they
measure nearly the same underlying property by different means, consistent
with the previous section, this property is closely tied to lexical
diversity (high correlation with RTTR). MOM and CorrInt are the most
effective for ranking generators, agreeing most with text-metric rankings
(Appendix~\ref{appendix_corr_results}). The remaining metrics correlate
little with each other. 


We then ask whether different testers rank generators consistently, computing
pairwise tester correlations per metric (Figure~\ref{fig:correclation_for_geom}).
Two patterns emerge. Tester \emph{capacity} matters: restricting to
high-capacity testers leaves the mean correlation unchanged but sharply
reduces its variance, so smaller testers give noisier rankings rather than
different ones. \emph{Task} matters too: agreement is substantially lower on
generation-from-title than on simplification or quantization.

\begin{figure}[hbt!]
    \centering
    \includegraphics[width=0.95\linewidth]{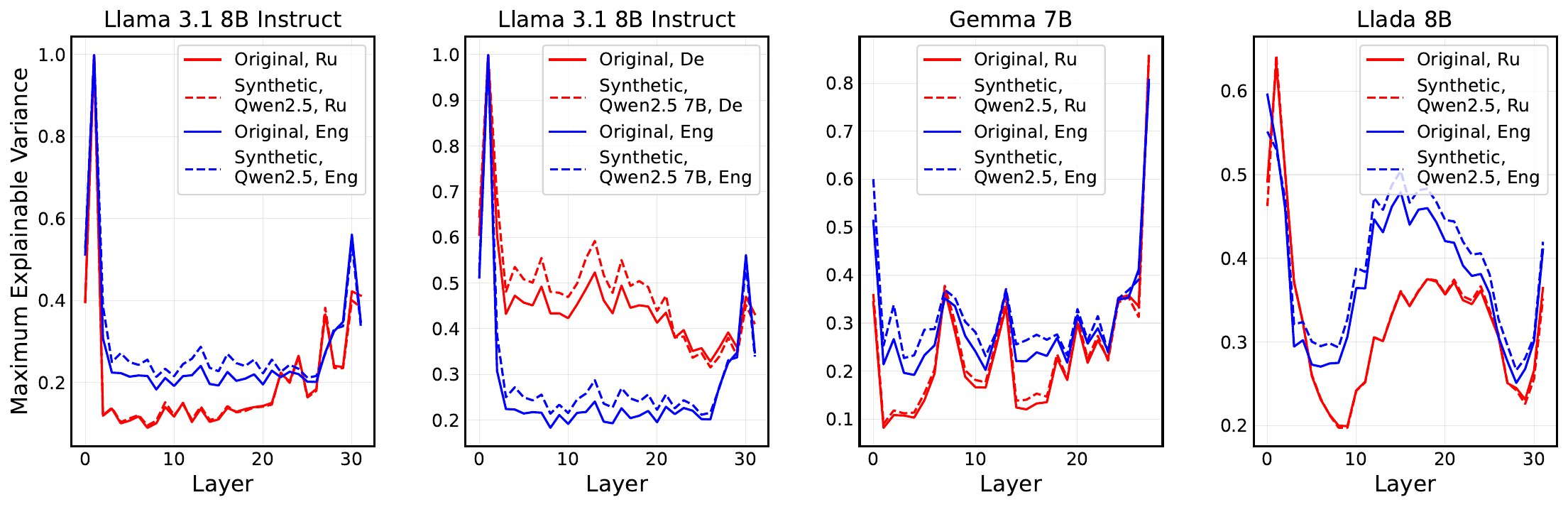}
    \caption{The comparison of Original and generated by Qwen2.5 7B synthetic texts on Russian (Ru), German (De) and English (Eng) for Maximum Explainable Variance \ref{eq:mev} and various tester models $\mathcal{T}$.}
    \label{fig:languages_main}
\end{figure}

\textbf{Consistency across testers and tasks.} The rewriting task
reinforces this picture. Despite spanning very different sizes,
architectures, and training paradigms (0.5B--8B; including the
diffusion-based LLaDA-8B), testers produce highly consistent generator
rankings, both layer-wise and on the layer average
(Figure~\ref{fig:layer_tot}).

\textbf{Cross-lingual behavior.} Finally, we vary the language of the
reviews (English, Russian, German; Figure~\ref{fig:languages_main}). The
metrics separate generated from original text clearly in English across all
testers, but the gap is much smaller in Russian plausibly because Russian
is under-represented in pretraining indicating that detectability degrades
for lower-resource languages.

\textbf{Quantization: Detectable Only When Degradation Is Severe.}
\label{sec:quant_main}
As a final stress-test we ask whether geometric metrics can detect
\emph{quality degradation induced by quantization}, a setting where the
generator is held fixed and only its numerical precision changes. We
evaluate LLaMA-3.1-8B and Mistral-7B at full precision (fp32), 4-bit
(AWQ), and 2-bit (AQLM), read by six testers
(setup in \S\ref{seq:datasets}).
\begin{figure}[t]
    \centering
    \includegraphics[width=0.97\linewidth]{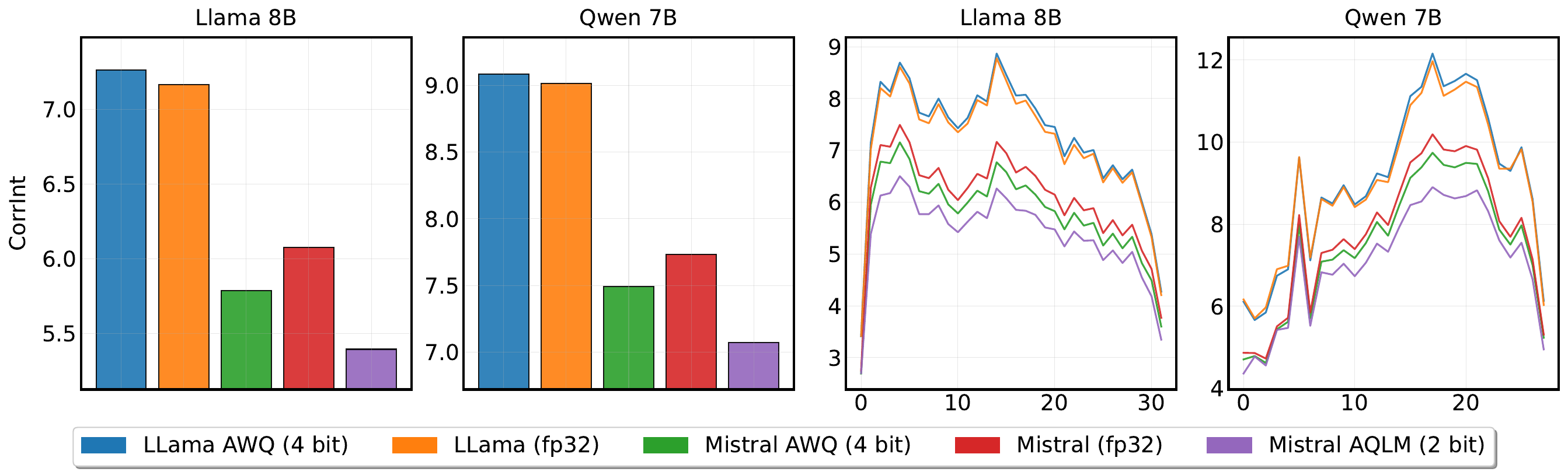}
    \caption{Average (left two columns) and layer-wise (right two columns)
    CorrInt for quantized generators, read by Qwen2-7B and Llama-3.1-8B
    testers. CorrInt drops for the 2-bit Mistral variant, where output
    quality is expected to be low, but does not cleanly separate the 4-bit
    LLaMA variant from full precision.}
    \label{fig:quant_main}
\end{figure}
The result is mixed, and we report it as such. CorrInt drops markedly for
the 2-bit (AQLM) Mistral variant the regime where generation quality is
expected to be most degraded so the metric does carry a signal for
\emph{severe} compression (Figure~\ref{fig:quant_main}). However, it does
\emph{not} cleanly distinguish the 4-bit (AWQ) LLaMA variant from full
precision, where the quality loss is mild. Geometric metrics therefore
detect quantization damage only once it is large enough to materially
change the generated text, and are not sensitive enough to flag the
moderate precision reductions most likely to be used in practice. Further
average, layer-wise, and correlation results across all six testers are
provided in Appendix~\ref{sec:quant_geometric_results}.

\section{Conclusion}

In this work, we present the first systematic stress-test of geometric metrics for LLM 
evaluation across six tester models ranging from 0.5B to 8B parameters, eight generator 
models, and multiple tasks and languages. We show that geometric metrics are genuinely 
useful for discriminative tasks  generator identification and human-vs-AI detection 
and that the intrinsic-dimensionality cluster selectively tracks lexical diversity, closely
correlated with RTTR.
However, they do not behave as general-purpose quality proxies: once length is controlled,
no metric shows broad, consistent alignment across our panel of established external text
metrics, and their signal 
depends critically on the choice and capacity of the tester model. We identify length 
confounding as a key methodological pitfall and recommend it be explicitly controlled in
future work. Taken together, our results offer a precise map of where geometric metrics 
can and cannot be trusted.

\clearpage

\bibliography{iclr2026_conference}
\bibliographystyle{iclr2026_conference}
\clearpage
\appendix
\section*{Appendix}


\textbf{The Appendix is organized as follows}:Section \ref{app:discussion} provides discussions, Section \ref{app:metrics} provides descriptions and definitions of geometric metrics, Section \ref{app:prompts} contains prompts for creating rewritten, generated from title and simplified texts, Sections \ref{appendix:text-metrics} describes utilized text metrics and provides the model evaluations with them. The Section \ref{sec:truthfulness} contains the Trustworthy analysis for Generator models $\mathcal{G}$, Sections \ref{sec:geometric_results}, \ref{sec:surface_similarity} and \ref{sec:perplexity} demonstrates the geometric, semantic, fluency and coherence analysis for Generator models across rewriting, Simplification and Generation from title tasks. The Sections \ref{sec:quant_geometric_results} and \ref{appendix} provides plots for quantization task and for other language tasks, respectively.

\section{Discussion and Limitations}
\label{app:discussion}
\vspace{-0.5em}

\textbf{Scoping what geometric metrics do.}
Prior work has demonstrated that geometric properties of LLM internal representations
show promise in isolated settings, but the conditions under which they are reliable,
and what they actually measure, had not been systematically investigated.
Our study provides that stress-test across a broad range of models, tasks, and conditions.
The picture that emerges is not negative but precise, geometric metrics are genuinely
useful for certain tasks, unreliable for others, and carry confounds that must be
explicitly managed.

\textbf{What geometric metrics are good for.}
Geometric metrics perform well in \emph{discriminative} settings.
On 6-class generator identification, adding all eight metrics to a text-statistics baseline
lifts accuracy from 69\% to 78\%; Schatten Norm alone contributes the largest individual
gain ($+5.2$ pp).
On human-vs-AI detection with a strong tester (Qwen2.5-7B), the full geometric feature
set reaches PR-AUC $= 0.95$, exceeding the text-statistics baseline of $0.89$.
The intrinsic dimensionality cluster (MLE, MOM, CorrInt, MADA) tracks lexical diversity
closely ($\rho \approx 0.54$ with RTTR on generated text), making these metrics
interpretable proxies for representational spread.

\textbf{What geometric metrics are not good for.}
General-purpose quality prediction is not a supported use case.
After length residualization, most partial Spearman correlations between geometric metrics
and our panel of established external text metrics (PPL, BLEURT, RTTR, Semantic Diversity,
Compression Rate) remain weak ($|\rho| < 0.3$), and no metric aligns consistently across the
panel. We frame this as convergent validity against specific, operationalized external
measures rather than against ``quality'' in the abstract, which the panel does not jointly
define. This is a new empirical finding rather than a rebuttal of
prior claims: it closes an open question about a plausible hypothesis, and the one robust
association we do find --- intrinsic dimensionality with lexical diversity --- matches
concurrent work~\citep{pedashenko2025unveiling}.

\textbf{The length confound.}
A key methodological finding is that text length is a pervasive confounder.
Schatten Norm and MOM lose the majority of their generator-discriminative signal after
length residualization. Since different generators produce systematically different output lengths,
any metric sensitive to length will appear discriminative without capturing geometric
properties of the representations. We recommend length residualization as a mandatory
diagnostic in any future work using these metrics.

\textbf{Dependence on the tester model.}
Geometric metrics must be used carefully because they do not measure text in isolation 
they reflect an interaction between the text and the tester model.
On human-written text, 88--97\% of the metric variance is attributable to the tester rather
than the paragraph itself, and consistent rankings require tester models of at least 7--8B
parameters. Beyond capacity, it is likely that the tester captures properties of the text
that go beyond length but whose nature remains unclear from our analysis alone.

\textbf{Limitations.}
Our analysis mainly focuses on English scientific abstracts and may not generalize to other genres.
Length residualization is linear, potentially leaving nonlinear length effects in residuals.
The 6-class identification task is a diagnostic stress-test rather than a deployment scenario.
Finally, the weak correlation between MLE and quality metrics reflects redundancy with the
specific quality dimensions we tested and may not exhaust the space of quality-relevant features.

\section{Description and Definition of Geometric Metrics}
\label{app:metrics}

The \textbf{Maximum Explainable Variance (MEV)} \citep{razzhigaev2024shape} and \textbf{Resultant Length} \citep{ethayarajh2019contextual} — often collectively referred to as measures of \textbf{Anisotropy} — quantify the degree of directional concentration in token embeddings. Lower values indicate more isotropic, evenly distributed representations, which are associated with greater semantic diversity and complexity. Empirically, generated text tends to exhibit higher MEV and Resultant Length compared to human text, making these metrics useful for distinguishing synthetic from natural content.

Let $X^{(l)} \in \mathbb{R}^{N \times d}$ denote the matrix of token representations from layer $l$, where each row corresponds to a token embedding. Let its singular value decomposition be:
\begin{equation}
X^{(l)} = U \Sigma V^\top, \quad \Sigma = \mathrm{diag}(\sigma_1, \sigma_2, \dots, \sigma_r), \quad r = \min(N, d),
\label{eq:svd}
\end{equation}
where $\sigma_1 \geq \sigma_2 \geq \dots \geq \sigma_r \geq 0$ are the singular values.

\paragraph{Maximum Explainable Variance (MEV).}
\begin{equation}
\mathrm{MEV}(X^{(l)}) = \frac{\sigma_1^2}{\sum_{i=1}^r \sigma_i^2}.
\label{eq:mev}
\end{equation}
This measures the proportion of total variance explained by the first principal component. High MEV indicates that representations are dominated by a single direction — a signature of representational collapse or anisotropy which demonstrates the diversity and complexity of the text.

\paragraph{Resultant Length ($R$).}
Let $\mathbf{x}_i = \frac{X^{(l)}_{i,:}}{\|X^{(l)}_{i,:}\|_2}$ be the unit-normalized $i$-th token embedding. Then:
\begin{equation}
R(X^{(l)}) = \left\| \frac{1}{N} \sum_{i=1}^N \mathbf{x}_i \right\|_2.
\label{eq:resultant}
\end{equation}
This measures the magnitude of the mean direction vector. $R = 0$ implies perfect isotropy; $R = 1$ implies perfect alignment. Like MEV, it is typically elevated in generated text to estimates its complexity.

\paragraph{Schatten Norm.}
The \textbf{Schatten-$p$ norm} \citep{bhatia1997matrix} provides a family of matrix norms that quantify global spectral energy in the representation matrix $X^{(l)}$. For $p \geq 1$, it is defined as:
\begin{equation}
\|X^{(l)}\|_{S_p} = \left( \sum_{i=1}^r \sigma_i^p \right)^{1/p},
\label{eq:schatten}
\end{equation}
where $\sigma_i$ are the singular values from the SVD in~\eqref{eq:svd}. The Schatten norm is not inherently better when higher or lower, its interpretation depends on the application. For instance, smaller nuclear norm may indicate more compact representations, while larger Frobenius norm may reflect higher activation energy or scale. It is often used for regularization, stability analysis, or comparing representation magnitudes across models or layers.

In contrast to the above, the \textbf{Effective Rank (ERank)} \cite{roy2007effective} quantifies the \textit{effective dimensionality} or \textit{diversity} of the representation space. It is a continuous, entropy-based approximation of matrix rank, robust to small perturbations.

Let $p_k = \frac{\sigma_k}{\sum_{i=1}^r \sigma_i}$ be variance proportion of the $k$-th singular value. Then:
\begin{equation}
\mathrm{ERank}(X^{(l)}) = \exp\left( -\sum_{k=1}^r p_k \log p_k \right).
\label{eq:erank}
\end{equation}
High ERank indicates that the model utilizes many orthogonal directions to encode information, which can be seen as a sign of diverse representations and the complex structure of layer outputs and texts.

\textbf{MAUVE} \citep{pillutla2021mauve} measures the divergence between generated and human text by comparing their distributions in a quantized embedding space. Let $P$ and $Q$ denote these distributions, quantized into a discrete space via clustering. Then:
\begin{equation}
\mathrm{MAUVE}(P, Q) = 1 - \inf_{\lambda \in [0,1]} \left[ \lambda \cdot D_{\mathrm{KL}}(P' \| R_\lambda) + (1 - \lambda) \cdot D_{\mathrm{KL}}(Q' \| R_\lambda) \right],
\label{eq:mauve}
\end{equation}
where $R_\lambda = \lambda P' + (1 - \lambda) Q'$. MAUVE produces a score between 0 and 1, where higher values indicate better alignment with human text. While designed for text, MAUVE can be applied to other modalities by using domain-specific embeddings and is best used to compare different models or decoding strategies rather than for absolute evaluation. We analyze texts and the internal representations $X_g^{(l)}$ mentioned above based on approach proposed in the original paper.

MAUVE has emerged as a prominent method for quantifying similarity between neural-generated and human-written text by computing divergence curves between their distributions, with higher scores indicating more coherent, human-like text \citep{xia2024ground, sen2025advancing}.

\paragraph{Intrinsic Dimensionality (ID).}
Intrinsic Dimensionality estimates the manifold dimension on which text representations lie. The idea of ID is to measure the minimum number of parameters required to represent the data without significant loss of information, which could be measured both locally and globally. Common estimators include the \textbf{Correlation Dimension} \citep{grassberger1983measuring}, the \textbf{Maximum Likelihood Estimator (MLE)} \citep{levina2004maximum} and others \cite{farahmand2007manifold, amsaleg2018extreme}. While not directly based on log-likelihood, lower ID often correlates with higher predictability, suggesting that the text lies on a simpler, more structured manifold. This allows to consider ID as an indicator of the model's ability to generalize.

\subsection{Quantization}
\label{sec:quant}

The main quantization result and its figure are presented in the main
text (\S\ref{sec:quant_main}, Figure~\ref{fig:quant_main}): CorrInt
detects severe (2-bit) degradation but does not cleanly separate mild
(4-bit) quantization from full precision. Full average, layer-wise, and
correlation results across all six tester models are provided in
Appendix~\ref{sec:quant_geometric_results}.

\section{Prompts for Text Generation and Rewriting}
\label{app:prompts}

The instructions provided to the models are defined as follows.

\begin{lstlisting}[caption={System prompt for movie reviews rewriting},label=lst:lay_summary]
Rewrite this text in a different style while preserving 
the main idea. Try to maintain the original length and language. 
Output only the rewritten text. Original text:''}
\end{lstlisting}

\begin{lstlisting}[caption={System prompt for generating lay summaries},label=lst:lay_summary2]
SYSTEM_PROMPT_LAY_SUMMARY = """You are a science communicator. Rewrite this abstract for a general audience by:
1. Defining any specialized terms in simple language
2. Focusing on why the findings matter to ordinary people
3. Using analogies or familiar concepts where helpful
4. Keeping the same core message but making it accessible

Output your summary only, nothing else.
Abstract: \n"""
\end{lstlisting}

\begin{lstlisting}[caption={System prompt for generating academic abstracts},label=lst:gen_abstract]
SYSTEM_PROMPT_GENERATE_ABS = """As a research scientist, write a plausible academic abstract based on the following paper title.
States the research problem or question. Clearly articulates the aim or hypothesis of the research.
Briefly describes the experimental design, data sources, models, or analytical approaches used.
Output your summary only, nothing else. Summarizes the main findings, often including key quantitative outcomes.
Avoid vague statements like "results are promising"; instead, give concrete outcomes. States the significance of the findings.
May mention broader impacts, applications, or future work.

Length: 150-300 words

Put it all into one paragraph, do not use specific sections, do not split text into dedicated paragraphs.

Title: \n"""
\end{lstlisting}

\section{Baseline Text Quality Metrics: Full Description}
\label{appendix:text-metrics}

We employ three reference-based and five reference-free metrics. 
All metrics are computed after light preprocessing: tokenization 
using NLTK's \texttt{word\_tokenize}, sentence segmentation via 
NLTK for semantic diversity, and optional filtering of scientific 
stopwords for RTTR. Semantic metrics use the 
\texttt{all-MiniLM-L6-v2} sentence transformer (6-layer, 384-dim 
embeddings).

\paragraph{Reference-Based Metrics.}

\textbf{ROUGE-L}~\citep{lin2004rouge} measures lexical overlap 
between generated and reference text via the longest common 
subsequence, capturing recall of content words without requiring 
exact n-gram matches.

\textbf{BLEURT}~\citep{sellam2020bleurt} is a learned metric based 
on BERT representations that correlates more strongly with human 
judgements than surface-level metrics, capturing semantic adequacy, 
entailment, and fluency.

\textbf{MAUVE}~\citep{pillutla2021mauve} measures distributional 
alignment between generated and human text by computing divergence 
curves in a quantized embedding space. Higher scores indicate 
closer distributional match to human text.

\paragraph{Reference-Free Metrics.}

\textbf{GPT-2 Perplexity (PPL)} measures fluency and linguistic 
coherence by computing how likely a fixed GPT-2 language model 
considers the generated text. Lower perplexity indicates more 
natural, fluent output.

\textbf{Compression Rate (CR)} estimates textual redundancy by 
compressing the text with a ZIP algorithm and computing the ratio 
of compressed to original length. Lower compression rate indicates 
higher lexical complexity and less repetition.

\textbf{Root Type-Token Ratio (RTTR)} measures lexical diversity 
as:
\[
    \mathrm{RTTR} = \frac{|\mathcal{V}|}{\sqrt{N}},
\]
where $|\mathcal{V}|$ is the number of unique tokens and $N$ is 
the total token count. Higher RTTR indicates richer vocabulary; 
very low values suggest oversimplification or repetition.

\textbf{Semantic Diversity (SemDiv)} captures conceptual variation 
across sentences by computing the average pairwise cosine distance 
between sentence embeddings:
\[
    \mathrm{SemDiv} = 1 - \frac{2}{n(n-1)} \sum_{i < j} 
    \cos(\mathbf{e}_i, \mathbf{e}_j),
\]
where $\mathbf{e}_i$ is the embedding of the $i$-th sentence and 
$n$ is the number of sentences.

\textbf{Flesch Reading Ease (FRE)} assesses accessibility via:
\[
    \mathrm{FRE} = 206.835 
    - 1.015 \cdot \frac{\text{words}}{\text{sentences}} 
    - 50.0 \cdot \frac{\text{syllables}}{\text{words}}.
\]
Higher scores indicate more readable text. We interpret FRE 
relative to domain norms, as scientific text is expected to score 
lower than general prose.

\subsection{LLM evaluation with text metrics}
\label{app:text_metrics_eval}
Comparison of text quality or naturalness metrics across original and generated texts for Movie reviews could be seen in Table \ref{tab:metrics_comparison}. Similar text metrics with analysis for Simplification and Generation from title tasks could be found in Appendices \ref{sec:surface_similarity} and \ref{sec:perplexity}. The study of Truthfulness for various Generator models $\mathcal{G}$ could be found in Appendices \ref{sec:truthfulness} and \ref{sec:truthfulness_results}. 

\begin{table}[htbp]
\centering
\caption{Comparison of text quality or naturalness metrics across original and generated texts.}
\label{tab:metrics_comparison}
\resizebox{\textwidth}{!}{%
\begin{tabular}{lccccccc}
\toprule
\textbf{Model} & 
{\textbf{CR}}  & 
{\textbf{ROUGE-L}} $(\uparrow)$ & 
{\textbf{BLEURT}} $(\uparrow)$& 
{\textbf{GPT-PPL} } $(\downarrow)$& 
{\textbf{MAUVE} } $(\uparrow)$& 
{\textbf{Avg Len}} & 
{\textbf{Std Len}} \\
\bottomrule
Original       & 0.5754 & {---} & 0.0000 & 42.35 & {---} & 18.25 & 7.03 \\
Gemma-2b-it    & 0.5957 & 0.3534 & -0.1764 & 31.10 & 0.01 & 17.33 & 3.26 \\
Qwen-2.5-7b-it       & 0.6148 & 0.4438 & -0.1499 & 38.51 & 0.06 & 16.18 & 3.71 \\
LLama-3.1-8B-it  & 0.6072 & 0.3028 & -0.3047 & 27.24 & 0.03 & 17.93 & 3.70 \\
Deepseek-R1   & 0.6093 & 0.4286 & -0.1108 & 56.08 & 0.15 & 17.05 & 11.51 \\
Mistral-7b-it     & 0.5903 & 0.3712 & -0.1663 & 35.57 & 0.11 & 16.75 & 3.70 \\
Phi-3-medium-4k-it     & 0.6018 & 0.3304 & -0.2599 & 47.28 & 0.11 & 17.32 & 4.08 \\
Phi-3-mini-128k-it    & 0.6182 & 0.2103 & -0.3561 & 49.32 & 0.02 & 22.22 & 5.13 \\
Starling-lm-7b-beta & 0.5791 & 0.4035 & -0.0904 & 30.47 & 0.01 & 19.22 & 3.68 \\
\hline
\end{tabular}%
 }
\end{table}

\subsection{Text length}
\label{app:text_length}
\begin{figure}[ht]
  \centering
  \includegraphics[width=0.9\linewidth]{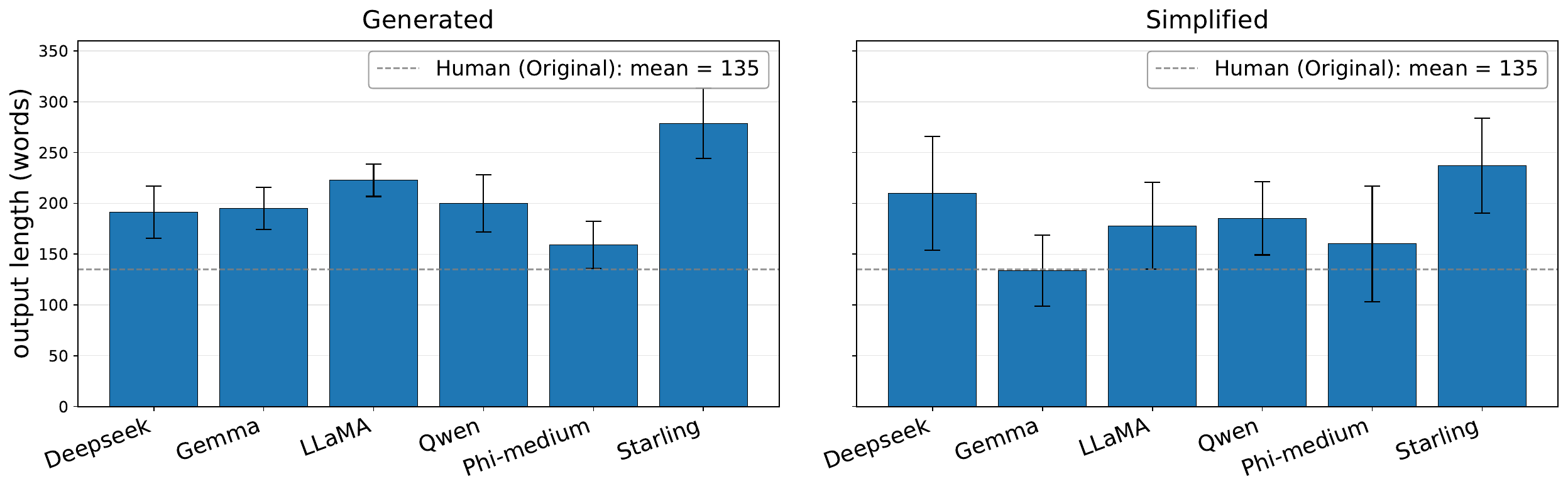}
  \caption{Output length per generator. Mean word count $\pm$ standard
  deviation. The dashed line marks the mean length of human abstracts
  (135 words). On \textsc{generated}, mean lengths span 159--279
  words, \textsc{starling} produces text roughly twice as long as
  \textsc{phi-medium}.}
  \label{fig:p1-lengths}
\end{figure}

The Figure \ref{fig:p1-lengths} demonstrates the generated text lengths in Generation and Simplification scenarios for $6$ Generator models $\mathcal{G}$. 

\subsection{Inter and Intra metrics Cross Correlation analysis}
\label{corr_method}
In this subsection, we analyze intra and inter correlation between text metrics and geometric metrics. To measure correlation between different metrics, we create the vectors of metric values for each of $N$ analyzed text for each metric. Then for these vectors the correlation is computed. More formally, let $m_k^i \in \mathbb{R}$ - the value of the metric $i$ for text $k$ and $M^i = [m_1^i, m_2^i, ..., m_N^i] \in \mathbb{R}^N$ the vector of metric $i$ value for all texts. Therefore, the correlation between metrics $i$ and $j$ could be computed as $C_{ij} = Spearman(M^i, M^j)$.
This correlation could be computed between text and geometric metrics as well as between metrics of the one group, when both metrics are text or geometric ones.

\subsection{Pseudocode}
\label{sec:pseudocode}

Algorithm~\ref{alg:scoring} outlines the scoring procedure applied to each model output.

\begin{algorithm}
\caption{Text Quality Scoring Pipeline}
\label{alg:scoring}
\begin{algorithmic}[1]
\Require Original abstract $A$, generated simplification $S$, generated title-based abstract $T$
\Ensure Metric scores for $A$, $S$, and $T$

\State Initialize \texttt{Scorer} with sentence transformer model
\For{each text $X \in \{A, S, T\}$}
    \If{$X$ is empty} 
        \State Set all metrics for $X$ to 0
        \State \textbf{continue}
    \EndIf
    
    \State $rttr_X \gets \Call{ComputeRTTR}{X, \text{scientific\_mode} = \text{False}}$
    \State $read_X \gets \Call{ComputeReadability}{X}$
    \State $semdiv_X \gets \Call{ComputeSemanticDiversity}{X, \text{use\_sentences} = \text{True}}$
\EndFor

\State Compute $\text{BERTScore}(A, S)$ using reference-target alignment
\State Write $(rttr_A, rttr_S, rttr_T, read_A, \dots, semdiv_T)$ to CSV
\end{algorithmic}
\end{algorithm}

\subsection{Implementation Notes}
\label{sec:implementation}

The full implementation uses Python with \texttt{sentence-transformers}, \texttt{nltk}, and \texttt{scikit-learn}. Key functions include:

\begin{lstlisting}[caption={RTTR Computation},label=lst:rttr]
def compute_rttr(text, scientific_mode=False):
    if invalid(text): return 0.0
    tokens = word_tokenize(text.lower())
    if scientific_mode:
        tokens = [t for t in tokens 
                  if t not in scientific_stopwords and len(t) > 2]
    if not tokens: return 0.0
    return len(set(tokens)) / sqrt(len(tokens))
\end{lstlisting}

\begin{lstlisting}[caption={Semantic Diversity via Sentence Embeddings},label=lst:semdiv]
def compute_semantic_diversity(text, use_sentences=True):
    if use_sentences:
        units = nltk.sent_tokenize(text)
        if len(units) < 2: return 0.0
    else:
        units = word_tokenize(text.lower())
    embeddings = model.encode(units)
    sims = cosine_similarity(embeddings)
    avg_sim = mean(upper_triangle(sims))
    return max(0, min(1, 1 - avg_sim))
\end{lstlisting}

\noindent The pipeline processes all model outputs in batch, writing results to a unified CSV for downstream analysis.

\subsection{Geometric and text metric correlations Results}
\label{appendix_corr_results}

The correlation between geometric metrics is presented in Figure~\ref{fig:geom_metrics_corr} for high-capacity 7-8B autoregressive models (Gemma 7B, LLama3.1 8B, Qwen2.5-7B) and only for the following metrics: MOM, Schatten Norm, CorrInt. The detailed variant with the mean and standard deviation values of Spearman correlation coefficients is presented in Appendix~\ref{appendix}, Figure~\ref{fig:geom_metrics_corr_annot}.

\begin{figure}
    \centering
    \includegraphics[width=1
    \linewidth]{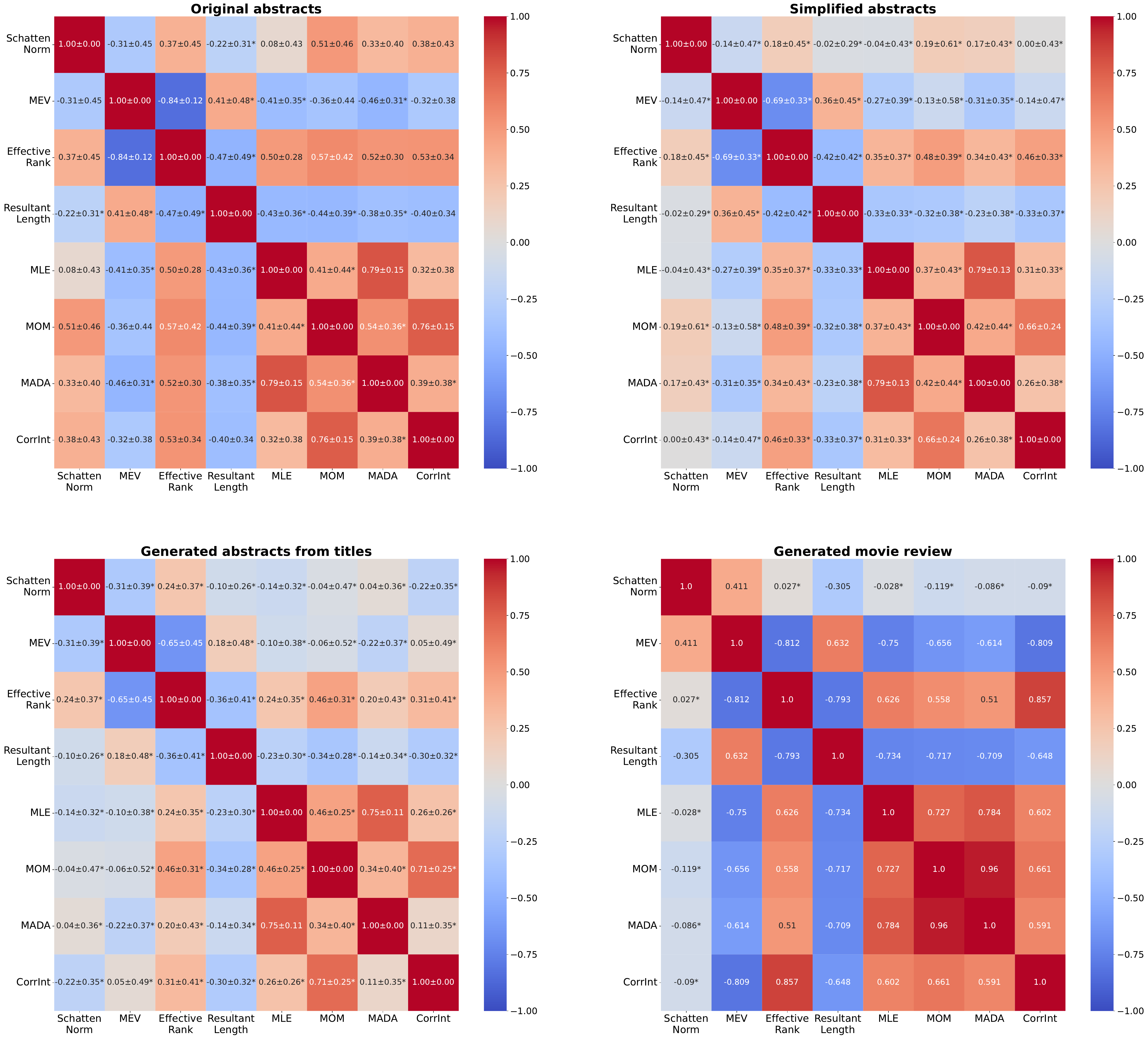}
    \caption{Geometric Metrics correlation with each other for four language tasks (Original abstracts, Simplified abstracts, Generated from titles and Rewritten movie reviews) for all Tester models $\mathcal{T}$. (With exact correlation values)}
    \label{fig:geom_metrics_corr_annot}
\end{figure}

\begin{table}[ht]
\centering
\caption{\textbf{Ranking:} Average Spearman correlation with other models for selected models and metrics across datasets.}
\small
\begin{tabular}{llcccccc}
\toprule
Dataset & Model & CorrInt &  MOM & Schatten Norm \\
\midrule
\multirow{3}{*}{Generated from Titles} & Gemma 7B & 0.49 &  0.69 & 0.20 \\
 & Llama3.1-8B & 0.44 &  0.48 & 0.30 \\
 & Qwen2.5-7B & 0.42 &  0.58 & 0.27 \\
\midrule
\multirow{3}{*}{Simplified Abstracts} & Gemma 7B & 0.53 &  0.73 & 0.36 \\
 & Llama3.1-8B & 0.49 &  0.66 & 0.39 \\
 & Qwen2.5-7B & 0.43 &  0.72 & 0.41 \\
 \midrule
\multirow{3}{*}{Quantized Models} & Gemma 7B & 0.89 & 0.77 & 0.53 \\
 & Qwen2.5-7B & 0.88 &  0.27 & 0.61 \\

\bottomrule
\end{tabular}

\label{tab:selected_correlations}
\end{table}

\begin{table}[ht]
\centering
\caption{Spearman correlation of RTTR with geometric metrics across datasets.}
\resizebox{\textwidth}{!}{%
\begin{tabular}{lcccccccc}
\toprule
Dataset & Schatten Norm & MEV & Effective Rank & Resultant Length & MLE & MOM & MADA & CorrInt \\
\midrule
Original abstracts & \textbf{0.65} & -0.09 & 0.36 & -0.29 & 0.38 & \textbf{0.77} & 0.45 & \textbf{0.71} \\
Generated from Titles & 0.15 & 0.18 & 0.18 & -0.20 & 0.43 & \textbf{0.60} & 0.25 & 0.34 \\
Simplified Abstracts & 0.43 & -0.01 & 0.30 & -0.16 & 0.32 & \textbf{0.51} & 0.29 & 0.47 \\
\bottomrule
\end{tabular}%
}

\label{tab:rttr_correlations}
\end{table}

\section{Truthfulness Evaluation}
\label{sec:truthfulness}

To assess the factual reliability of generated responses, we evaluate models on two established question-answering benchmarks: \textbf{SimpleQA}~\cite{simpleqa} and \textbf{TruthfulQA}~\cite{lin2022truthfulqa}. Our pipeline combines automated generation with LLM-as-judge verification using DeepSeek-V3.

\subsection{Datasets}
\begin{itemize}[leftmargin=*]
    \item \textbf{SimpleQA} is a dataset of factual questions paired with single-sentence ground-truth answers (e.g., ``What is the capital of France?'' → ``Paris''). It tests basic factual recall and precision.
    
    \item \textbf{TruthfulQA} contains adversarial questions designed to elicit false or misleading responses from language models (e.g., ``What happens if you swallow gum?''). Each question is annotated with a human-written ``best answer'' that is both truthful and informative. The dataset explicitly measures a model’s tendency to generate plausible but incorrect statements.
\end{itemize}

\subsection{Evaluation Strategy}
We adopt a two-stage protocol:

\begin{enumerate}[leftmargin=*]
    \item \textbf{Generation}: Models are prompted with a system instruction emphasizing truthfulness and conciseness:
    \begin{quote}
    \texttt{You are a helpful and truthful AI assistant. Answer the following question concisely and accurately. Provide only the answer, without any extra text.}
    \end{quote}
    Responses are generated using Ollama-hosted models at temperature 0.0 to minimize stochasticity.

    \item \textbf{Judgment}: A DeepSeek-V3 judge evaluates each response:
    \begin{itemize}
        \item For \textbf{SimpleQA}, the judge performs fact-checking against the reference answer. If an exact normalized match fails, it assesses factual correctness via a structured prompt and returns a binary verdict (\texttt{is\_correct}) with justification.
        
        \item For \textbf{TruthfulQA}, the judge evaluates two dimensions independently: (1) \textit{truthfulness} (absence of false/misleading claims) and (2) \textit{informativeness} (relevance and utility), following the original evaluation protocol~\cite{lin2022truthfulqa}.
    \end{itemize}
    All judgments are elicited via constrained JSON output and parsed robustly to handle formatting inconsistencies.
\end{enumerate}

This hybrid approach—combining exact matching where appropriate and LLM-based semantic verification otherwise—enables scalable yet nuanced truthfulness assessment across diverse factual domains.

\section{Truthfulness Evaluation Results}
\label{sec:truthfulness_results}

We evaluated seven open-weight language models on factual reliability using two established benchmarks: \textbf{SimpleQA} (factual recall) and \textbf{TruthfulQA} (resistance to generating false or misleading statements). For each model, we generated responses to all questions in both datasets using a standardized truthfulness-focused prompt (temperature = 0.0). Responses were then assessed by a DeepSeek-V3 judge using dataset-specific protocols:

\begin{itemize}[leftmargin=*]
    \item On \textbf{SimpleQA}, correctness was determined first by exact normalized string match; non-matching answers were judged for factual accuracy via structured LLM evaluation, yielding a binary \texttt{is\_correct} label.
    \item On \textbf{TruthfulQA}, the judge independently rated each response for \textit{truthfulness} (absence of falsehoods) and \textit{informativeness} (relevance and utility), following the original evaluation design~\cite{lin2022truthfulqa}.
\end{itemize}

Final metrics represent the percentage of correct or positively rated responses across all questions in each dataset. Results are summarized in Table~\ref{tab:truthfulness_results}.

\begin{table}[ht]
\centering
\caption{Truthfulness evaluation results across models and datasets. All metrics are reported as percentages. \textbf{Higher is better} for all columns.}
\label{tab:truthfulness_results}
\resizebox{\columnwidth}{!}{%
\begin{tabular}{lrrr}
\toprule
\textbf{Model} & 
\textbf{SimpleQA Accuracy} {\small (↑ higher is better)} & 
\textbf{TruthfulQA Truthful} {\small (↑ higher is better)} & 
\textbf{TruthfulQA Informative} {\small (↑ higher is better)} \\
\midrule
\texttt{deepseek\_r1\_8b}             & 3.95\%  & 70.13\% & 63.16\% \\
\texttt{llama3\_1\_8b\_instruct}      & 6.75\%  & 70.76\% & 75.06\% \\
\texttt{mistral\_7b\_instruct}        & 5.27\%  & 79.37\% & 82.03\% \\
\texttt{phi3\_medium\_4k\_instruct}   & 5.80\%  & 72.28\% & 63.54\% \\
\texttt{phi3\_mini\_128k\_instruct}   & 1.80\%  & 4.94\%  & 3.54\%  \\
\texttt{qwen2\_5\_7b\_instruct}       & 4.05\%  & 67.85\% & 52.03\% \\
\texttt{starling\_lm\_7b\_beta}       & 5.34\%  & 67.72\% & 85.70\% \\
\bottomrule
\end{tabular}%
}
\end{table}

\section{Geometric and Semantic Quality of Generated Texts}
\label{sec:geometric_results}

We evaluate the linguistic and semantic properties of model-generated scientific texts across two tasks: (1) \textit{abstract generation and simplification}, and (2) \textit{review rewriting}. Metrics include lexical diversity (Root Type-Token Ratio, RTTR), readability (Flesch Reading Ease), and semantic diversity (based on sentence embeddings). Higher RTTR and semantic diversity indicate richer vocabulary and conceptual variation, while higher readability scores reflect greater accessibility.

Critically, all models process the **same original texts**, so metrics for original abstracts and reviews are identical across models. Differences arise only in generated outputs.

\begin{table}[ht]
\centering
\caption{Average metrics for generated scientific abstracts. Original abstract metrics are shared across all models (leftmost column). \textbf{Title - means generated from title.}}
\label{tab:abstract_metrics}
\resizebox{\columnwidth}{!}{%
\begin{tabular}{lrrrrrrrrr}
\toprule
\multirow{2}{*}{\textbf{Model}} & 
\multicolumn{1}{c}{\textbf{RTTR}} & 
\multicolumn{2}{c}{\textbf{RTTR}} & 
\multicolumn{1}{c}{\textbf{Readability}} & 
\multicolumn{2}{c}{\textbf{Readability}} & 
\multicolumn{1}{c}{\textbf{SemDiv}} & 
\multicolumn{2}{c}{\textbf{SemDiv}} \\
\cmidrule(lr){2-2} \cmidrule(lr){3-4} \cmidrule(lr){5-5} \cmidrule(lr){6-7} \cmidrule(lr){8-8} \cmidrule(lr){9-10}
& orig & title & simple & orig & title & simple & orig & title & simple \\
\midrule
\texttt{deepseek\_r1\_8b}             & \multirow{8}{*}{7.111} & 8.78 & 8.73 & \multirow{8}{*}{95.991} & 81.90 & 99.83 & \multirow{8}{*}{0.575} & 0.507 & 0.629 \\
\texttt{gemma2\_2b\_it}               & & 8.64 & 7.90 & & 86.16 & 99.98 & & 0.538 & 0.593 \\
\texttt{llama3\_1\_8b\_instruct}      & & 8.63 & 8.74 & & 88.31 & 99.87 & & 0.518 & 0.624 \\
\texttt{mistral\_7b\_instruct}        & & 8.64 & 8.93 & & 88.45 & 99.64 & & 0.537 & 0.650 \\
\texttt{phi3\_medium\_4k\_instruct}   & & 9.19 & 9.13 & & 81.96 & 99.68 & & 0.559 & 0.615 \\
\texttt{phi3\_mini\_128k\_instruct}   & & 6.15 & 4.78 & & 98.56 & 89.15 & & 0.667 & 0.448 \\
\texttt{qwen2\_5\_7b\_instruct}       & & 9.30 & 8.91 & & 85.84 & 99.97 & & 0.564 & 0.627 \\
\texttt{starling\_lm\_7b\_beta}       & & 8.76 & 8.77 & & 85.46 & 98.21 & & 0.505 & 0.622 \\
\bottomrule
\end{tabular}%
}
\end{table}

\begin{table}[ht]
\centering
\caption{Average metrics for generated movie reviews. Original review metrics are shared across all models.}
\label{tab:review_metrics}
\resizebox{\columnwidth}{!}{%
\begin{tabular}{lrrrrrr}
\toprule
\textbf{Model} & 
\textbf{RTTR\_orig} & \textbf{RTTR\_gen} & 
\textbf{Readability\_orig} & \textbf{Readability\_gen} & 
\textbf{SemDiv\_orig} & \textbf{SemDiv\_gen} \\
\midrule
\texttt{rewrite\_deepseek\_eng}           & \multirow{8}{*}{8.822} & 9.72 & \multirow{8}{*}{97.744} & 98.98 & \multirow{8}{*}{0.733} & 0.733 \\
\texttt{rewrite\_gemma-2-baku-2b-it\_eng} & & 9.46 & & 99.88 & & 0.718 \\
\texttt{rewrite\_llama31\_eng}            & & 9.04 & & 99.96 & & 0.727 \\
\texttt{rewrite\_mistral\_7b\_it\_eng}    & & 9.09 & & 98.93 & & 0.724 \\
\texttt{rewrite\_phi-3-medium-4k\_eng}    & & 9.35 & & 99.46 & & 0.698 \\
\texttt{rewrite\_phi3-mini\_eng}          & & 8.78 & & 97.70 & & 0.674 \\
\texttt{rewrite\_qwen25\_eng}             & & 9.10 & & 99.89 & & 0.733 \\
\texttt{rewrite\_starling-lm-7b-beta\_eng}& & 9.33 & & 99.81 & & 0.703 \\
\bottomrule
\end{tabular}%
}
\end{table}

\paragraph{Key Observations.}
\begin{itemize}[leftmargin=*]
    \item The \texttt{orig} columns are identical across models, confirming evaluation on a shared dataset.
    
    \item \textbf{Readability inflation}: Generated abstracts achieve near-perfect readability ($\geq$98), suggesting models favor short, simple sentences—potentially sacrificing technical nuance.
    
    \item \textbf{Lexical diversity trade-offs}: Most models increase RTTR in generated text, but \texttt{phi3\_mini\_128k\_instruct} shows a sharp drop in RTTR for simplified abstracts (4.78 vs. 7.11), indicating excessive repetition or loss of content.
    
    \item \textbf{Semantic coherence}: In abstract simplification, semantic diversity generally increases from title-based to simplified outputs (e.g., \texttt{mistral\_7b}: 0.537 → 0.650), suggesting added explanatory content. However, \texttt{phi3\_mini} shows a reversal (0.667 → 0.448), signaling degraded simplification quality.
    
    \item \textbf{Review fidelity}: Review rewrites maintain high semantic similarity to originals (SemDiv $\approx$ 0.70–0.73), with most models preserving or slightly adjusting lexical diversity.
\end{itemize}

These results underscore that while larger models balance simplicity and richness effectively, smaller architectures like \texttt{phi3\_mini} struggle with scientific text simplification without compromising lexical or semantic quality.

Notably, all models exhibit low accuracy on SimpleQA ($\leq$6.75\%), suggesting challenges in precise factual recall without retrieval augmentation. In contrast, most models achieve substantially higher truthfulness rates on TruthfulQA (67-79\%), indicating better capability to avoid overt falsehoods when answering adversarial questions—though informativeness varies widely (e.g., \texttt{starling\_lm\_7b\_beta} excels in informativeness despite modest truthfulness). The \texttt{phi3\_mini\_128k\_instruct} model underperforms dramatically on TruthfulQA, possibly due to insufficient training on scientific or encyclopedic knowledge.

\section{Surface-Level Similarity in Abstract Generation and Simplification}
\label{sec:surface_similarity}

To assess surface-level fidelity between model outputs and reference texts, we compute \textbf{ROUGE-L} (longest common subsequence) and \textbf{BLEURT} (learned semantic similarity) scores for both tasks:
\begin{itemize}[leftmargin=*]
    \item \textit{Simplification}: comparison of generated simplified abstracts against original abstracts.
    \item \textit{Title-based generation}: comparison of abstracts generated from paper titles against original abstracts.
\end{itemize}
Higher scores indicate greater overlap or semantic alignment with the reference. Results are averaged per model and reported in Table~\ref{tab:surface_metrics}.

\begin{table}[ht]
\centering
\caption{ROUGE-L and BLEURT scores for abstract simplification and title-based generation. All scores are averaged over the evaluation set. Higher is better.}
\label{tab:surface_metrics}
\small 
\renewcommand{\arraystretch}{1.3}
\begin{tabular}{lrrrr}
\toprule
\textbf{Model} & 
\multicolumn{1}{c}{\rotatebox{60}{\textbf{ROUGE-L (simple)}}} & 
\multicolumn{1}{c}{\rotatebox{60}{\textbf{BLEURT (simple)}}} & 
\multicolumn{1}{c}{\rotatebox{60}{\textbf{ROUGE-L (title)}}} & 
\multicolumn{1}{c}{\rotatebox{60}{\textbf{BLEURT (title)}}} \\
\midrule
\texttt{deepseek\_r1\_8b}             & 0.343 & 0.189 & 0.377 & 0.169 \\
\texttt{gemma2\_2b\_it}               & 0.358 & 0.156 & 0.415 & 0.167 \\
\texttt{llama3\_1\_8b\_instruct}      & 0.357 & 0.163 & 0.396 & 0.172 \\
\texttt{mistral\_7b\_instruct}        & 0.377 & 0.239 & 0.390 & 0.167 \\
\texttt{phi3\_medium\_4k\_instruct}   & 0.345 & 0.162 & 0.382 & 0.158 \\
\texttt{phi3\_mini\_128k\_instruct}   & 0.135 & 0.071 & 0.148 & 0.093 \\
\texttt{qwen2\_5\_7b\_instruct}       & 0.365 & 0.177 & 0.392 & 0.164 \\
\texttt{starling\_lm\_7b\_beta}       & \textbf{0.428} & \textbf{0.284} & 0.398 & 0.164 \\
\bottomrule
\end{tabular}
\end{table}

\paragraph{Key Observations.}
\begin{itemize}[leftmargin=*]
    \item \textbf{Starling-LM-7B excels at simplification}: It achieves the highest ROUGE-L (0.428) and BLEURT (0.284) for simplified abstracts, suggesting it best preserves content while rephrasing for accessibility.
    
    \item \textbf{Title-based generation is more consistent}: ROUGE-L scores for title-generated abstracts are generally higher and less variable (0.377–0.415) than for simplification, indicating models find it easier to generate plausible abstracts from titles than to faithfully simplify existing ones.
    
    \item \textbf{Phi-3-mini underperforms severely}: With ROUGE-L below 0.15 in both tasks, it fails to retain core content—consistent with its poor performance in geometric and truthfulness evaluations.
    
    \item \textbf{BLEURT vs. ROUGE-L divergence}: Mistral-7B shows high BLEURT (0.239) but moderate ROUGE-L (0.377) in simplification, suggesting it captures semantic meaning despite surface-level differences. Conversely, Gemma2 achieves higher ROUGE-L but lower BLEURT, indicating lexical overlap without deep semantic alignment.
\end{itemize}

These results highlight a critical trade-off: while some models produce fluent and readable outputs (as seen in Section~\ref{sec:geometric_results}), they may not preserve factual or semantic content—underscoring the need for multi-faceted evaluation beyond surface metrics.

\section{Fluency and Coherence via Perplexity}
\label{sec:perplexity}

To evaluate the fluency and linguistic coherence of generated texts, we compute **perplexity (PPL)** using a fixed GPT-2 language model as a reference. Lower perplexity indicates that the text is more probable under a standard language model—i.e., more fluent and syntactically natural. We report PPL for both simplified abstracts (`PPL-simple`) and title-generated abstracts (`PPL-title`), with results averaged per model in Table~\ref{tab:perplexity}.

\begin{table}[ht]
\centering
\caption{GPT-2 perplexity (PPL) of generated abstracts. Lower values indicate higher fluency. All models are evaluated on the same set of outputs using the same GPT-2 checkpoint.}
\label{tab:perplexity}
\begin{tabular}{lrr}
\toprule
\textbf{Model} & \textbf{PPL (simple)} & \textbf{PPL (title)} \\
\midrule
\texttt{deepseek\_r1\_8b}             & 27.40 & 37.47 \\
\texttt{gemma2\_2b\_it}               & 19.62 & 19.86 \\
\texttt{llama3\_1\_8b\_instruct}      & 16.06 & 17.76 \\
\texttt{mistral\_7b\_instruct}        & 23.72 & 17.78 \\
\texttt{phi3\_medium\_4k\_instruct}   & 31.49 & 35.77 \\
\texttt{phi3\_mini\_128k\_instruct}   & 640.10 & 290.54 \\
\texttt{qwen2\_5\_7b\_instruct}       & 21.62 & 25.30 \\
\texttt{starling\_lm\_7b\_beta}       & 20.83 & \textbf{14.40} \\
\bottomrule
\end{tabular}
\end{table}

\paragraph{Key Observations.}
\begin{itemize}[leftmargin=*]
    \item \textbf{Llama-3-8B achieves highest fluency in simplification}: With the lowest PPL (16.06) on simplified abstracts, it produces the most GPT-2–plausible outputs, suggesting strong control over scientific language.
    
    \item \textbf{Starling-LM excels in title-based generation}: It attains the lowest PPL (14.40) for title-generated abstracts, indicating highly coherent and fluent synthesis from minimal prompts.
    
    \item \textbf{Phi-3-mini shows catastrophic degradation}: Its PPL exceeds 600 for simplification—orders of magnitude higher than other models—revealing severe incoherence, likely due to insufficient training on technical or long-form scientific text.
    
    \item \textbf{Task asymmetry}: Most models exhibit lower PPL on simplified abstracts than on title-generated ones (e.g., Llama-3: 16.06 vs. 17.76), suggesting that rewriting existing content is easier than generating from scratch. Notably, Starling-LM reverses this trend, excelling at generation.
\end{itemize}

These perplexity results align with our geometric and truthfulness evaluations: models that perform well across multiple axes (e.g., Llama-3, Starling-LM) also produce more fluent text, while smaller models like Phi-3-mini fail consistently across all metrics.

\section{Results for Qunatized models}
\label{sec:quant_geometric_results}

\begin{figure}[ht]
    \centering
    \includegraphics[width=1\linewidth]{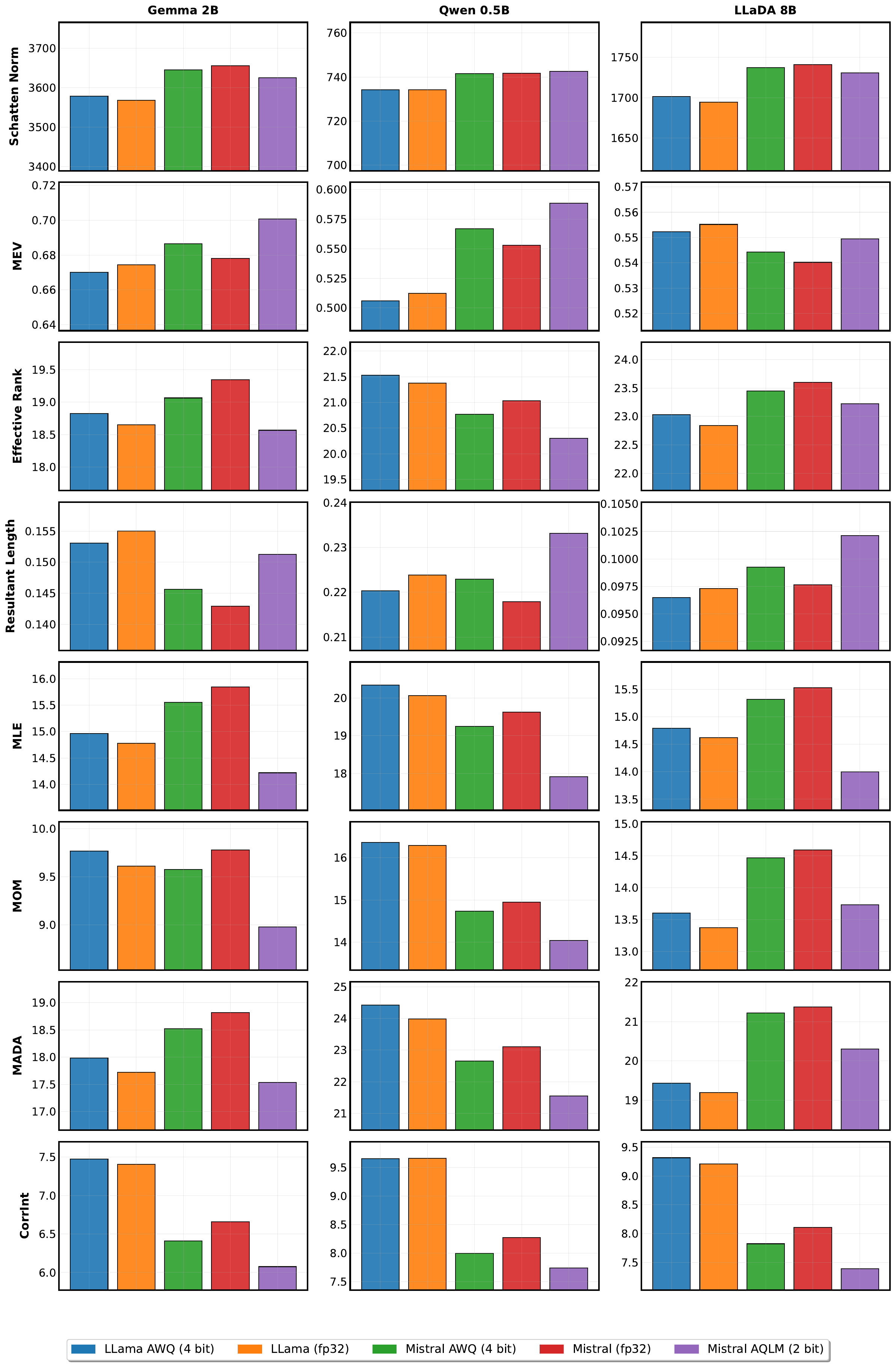}
    \caption{The ranking of two generators models with different quantization techniques via tester models (Qwen2 0.5B, Gemma 2B and LLaDA 8B) and all geometric metrics.}
    \label{fig:quantized_total_1}
\end{figure}

\begin{figure}[ht]
    \centering
    \includegraphics[width=1\linewidth]{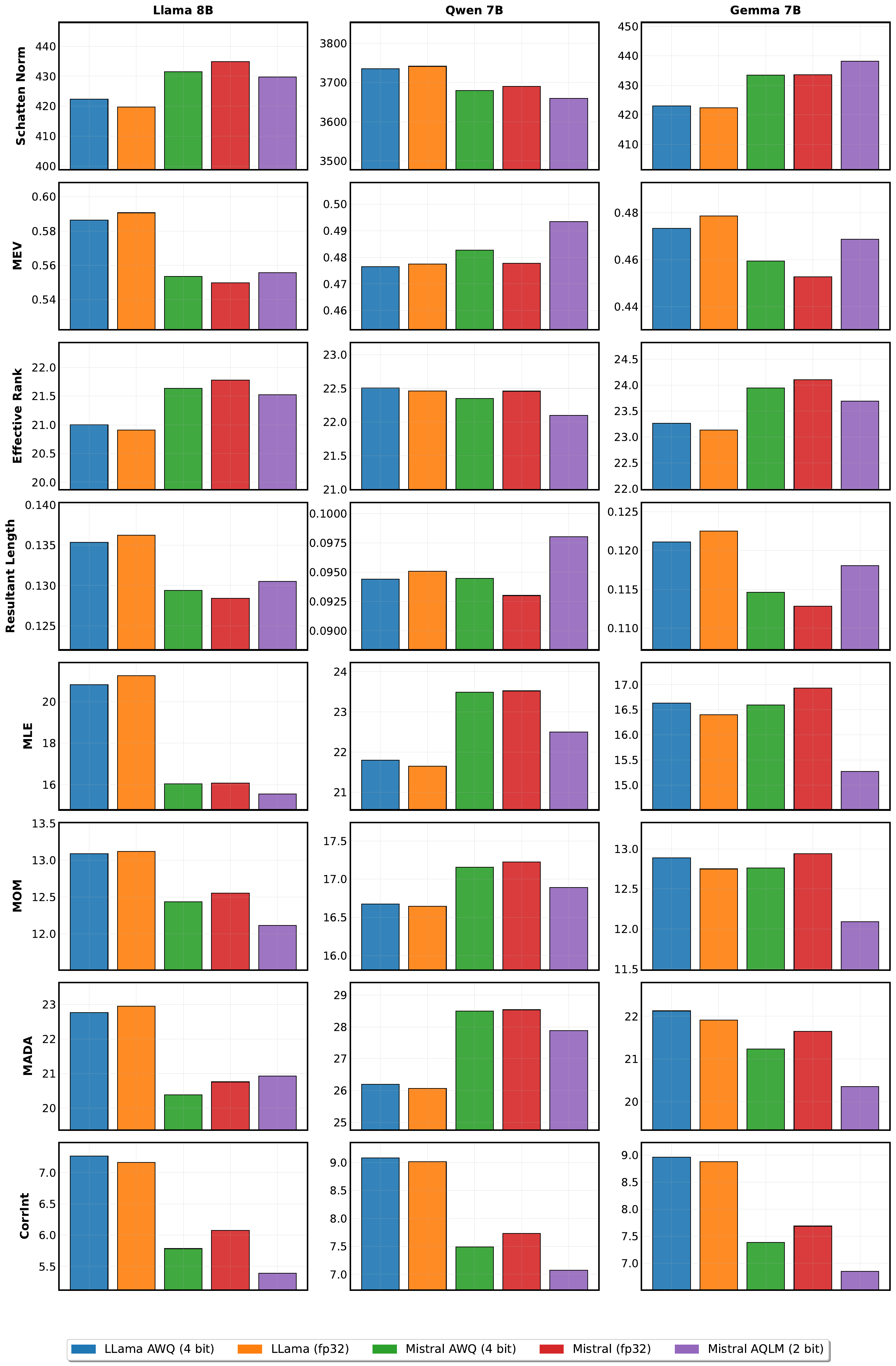}
    \caption{The ranking of two generators models with different quantization techniques via tester models (Qwen2.5 7B Instruct, Gemma 7B and Llama3.1 8B Instruct) and all geometric metrics.}
    \label{fig:quantized_total_2}
\end{figure}

\begin{figure}[ht]
    \centering
    \includegraphics[width=1\linewidth]{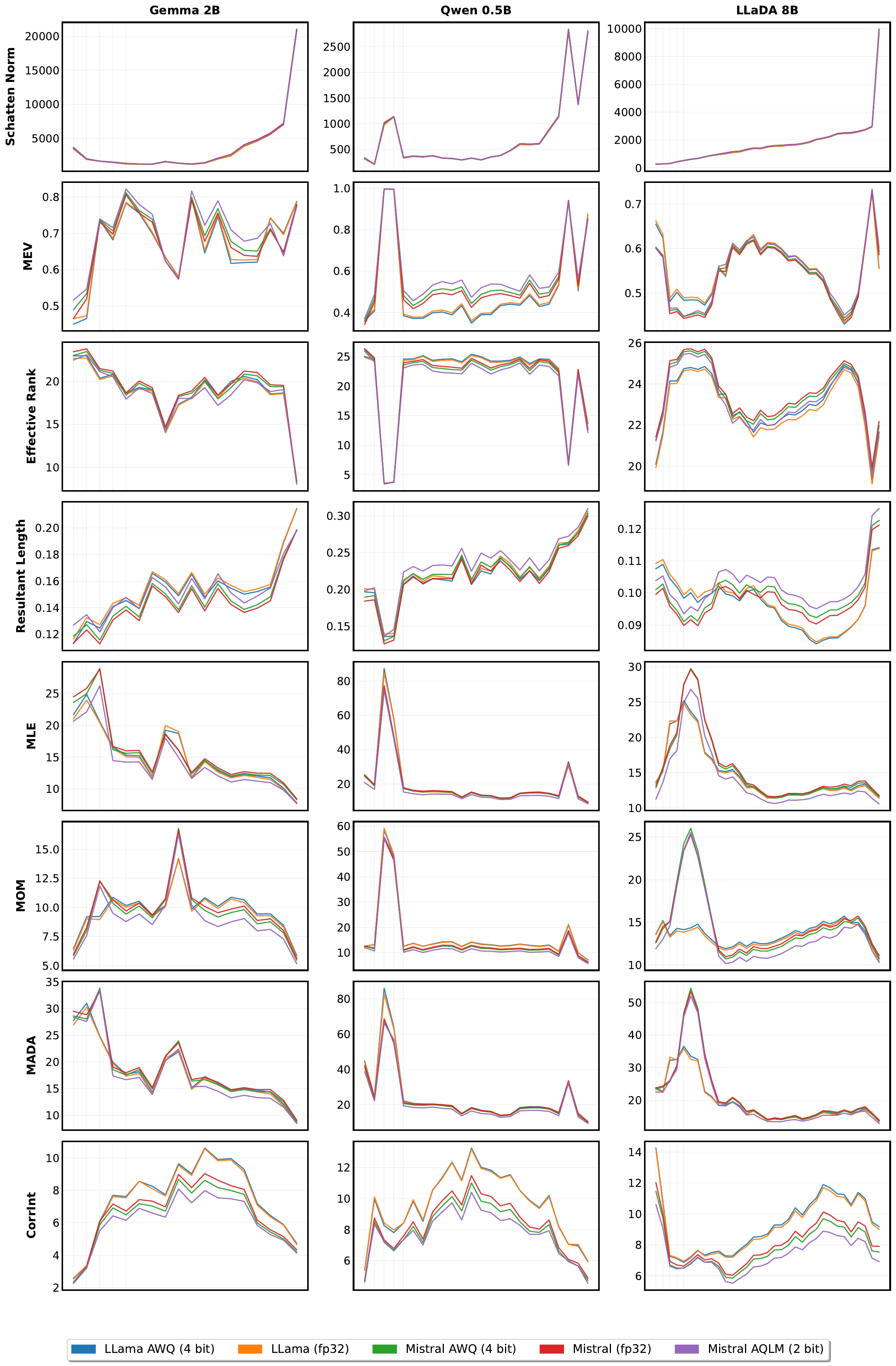}
    \caption{The layer-wize ranking of two generators models with different quantization techniques via tester models (Qwen2 0.5B, Gemma 2B and LLaDA 8B) and all geometric metrics.}
    \label{fig:quantized_layer_1}
\end{figure}

\begin{figure}[ht]
    \centering
    \includegraphics[width=1\linewidth]{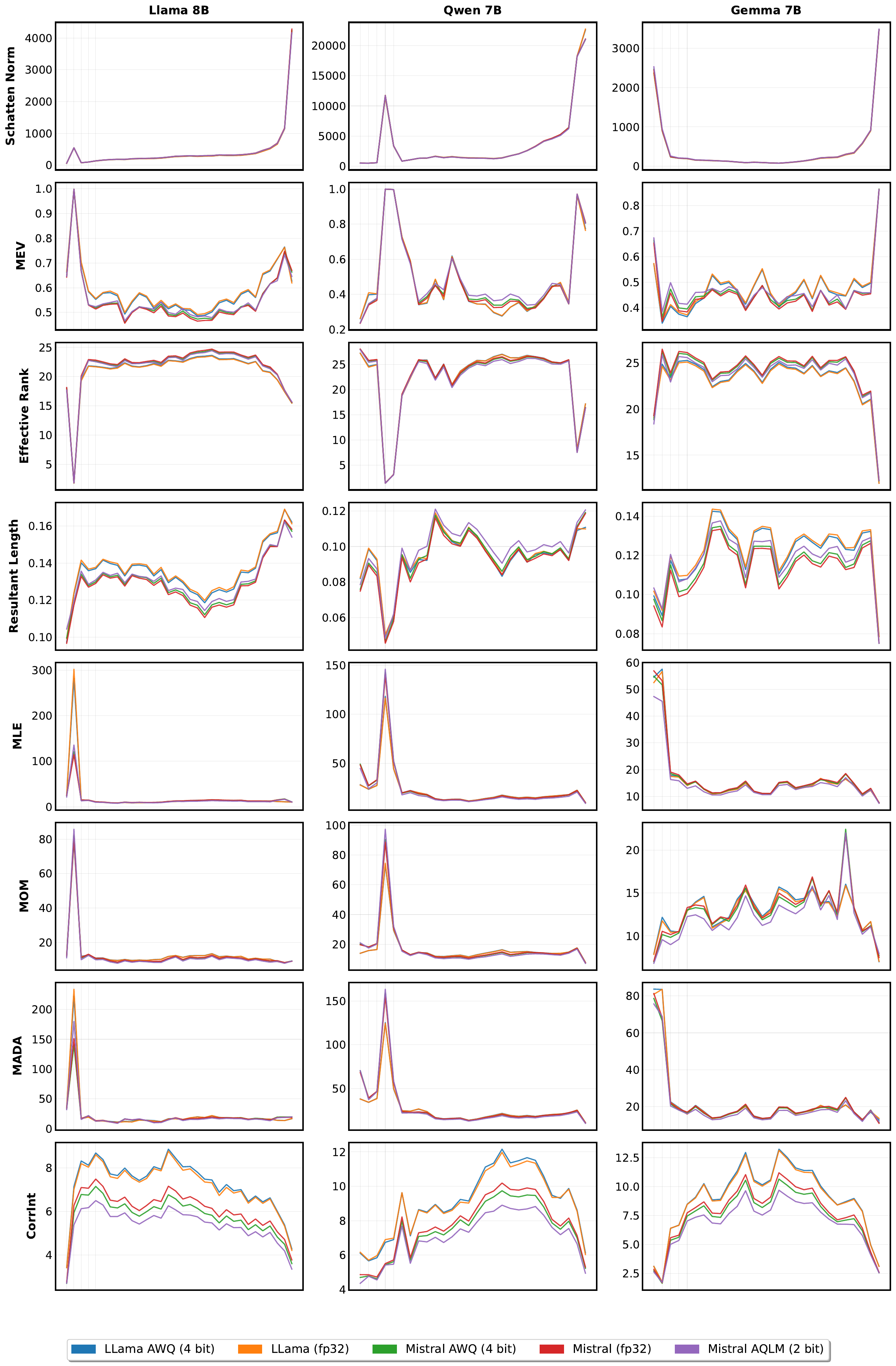}
    \caption{The layer-wize ranking of two generators models with different quantization techniques via tester models (Qwen2.5 7B Instruct, Gemma 7B and Llama3.1 8B Instruct) and all geometric metrics.}
    \label{fig:quantized_layer_2}
\end{figure}

\begin{figure}[ht]
\centering
    \includegraphics[width=0.9\linewidth]{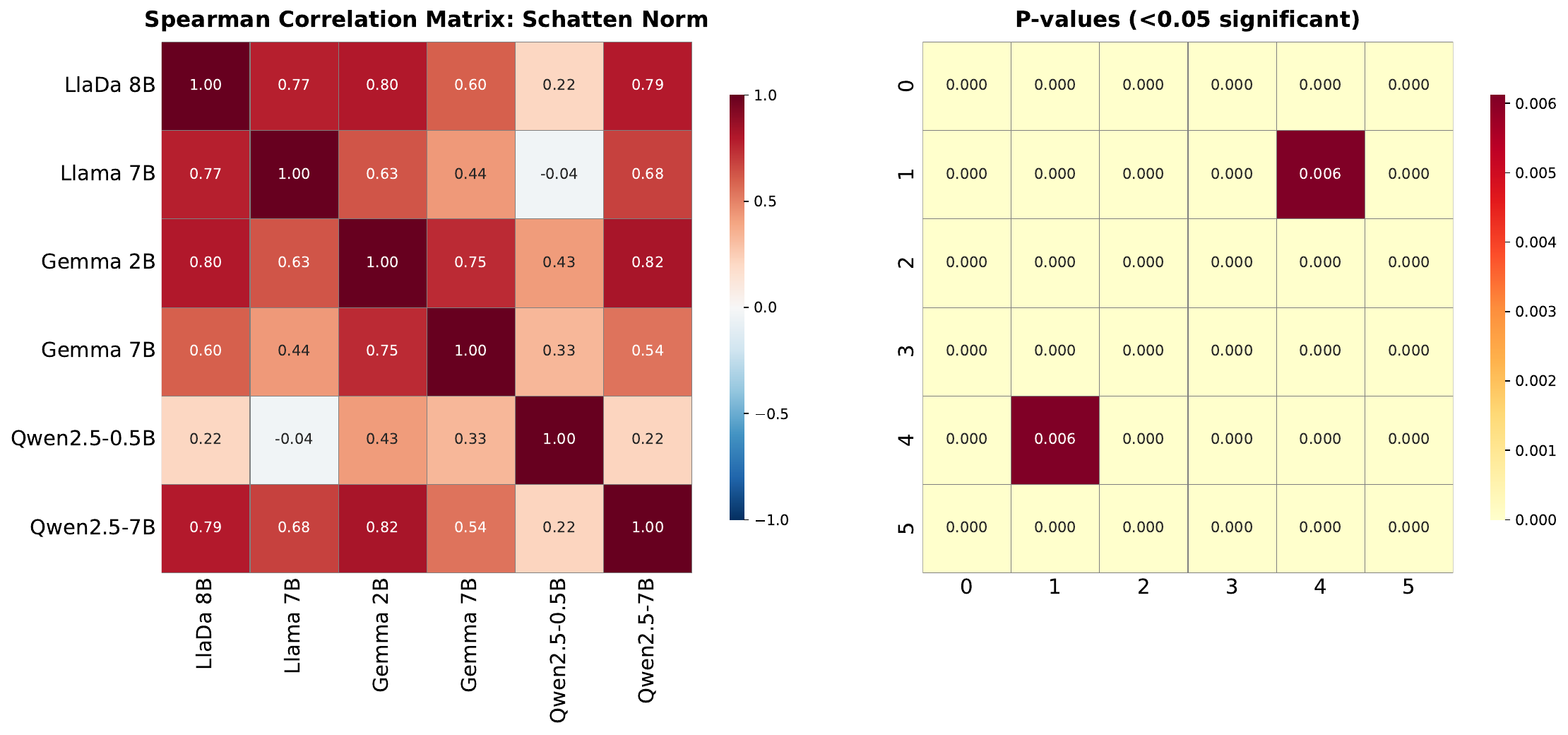}
    \caption{The correlation analysis of quantized models for the following generator models: LLama AWQ (4 bit), LLama (fp32), Mistral AWQ (4 bit), Mistral (fp32), Mistral AQLM (2 bit)}
    \label{fig:cor_quan_Schatten Norm}
\end{figure}

\begin{figure}[ht]
\centering
    \includegraphics[width=0.9\linewidth]{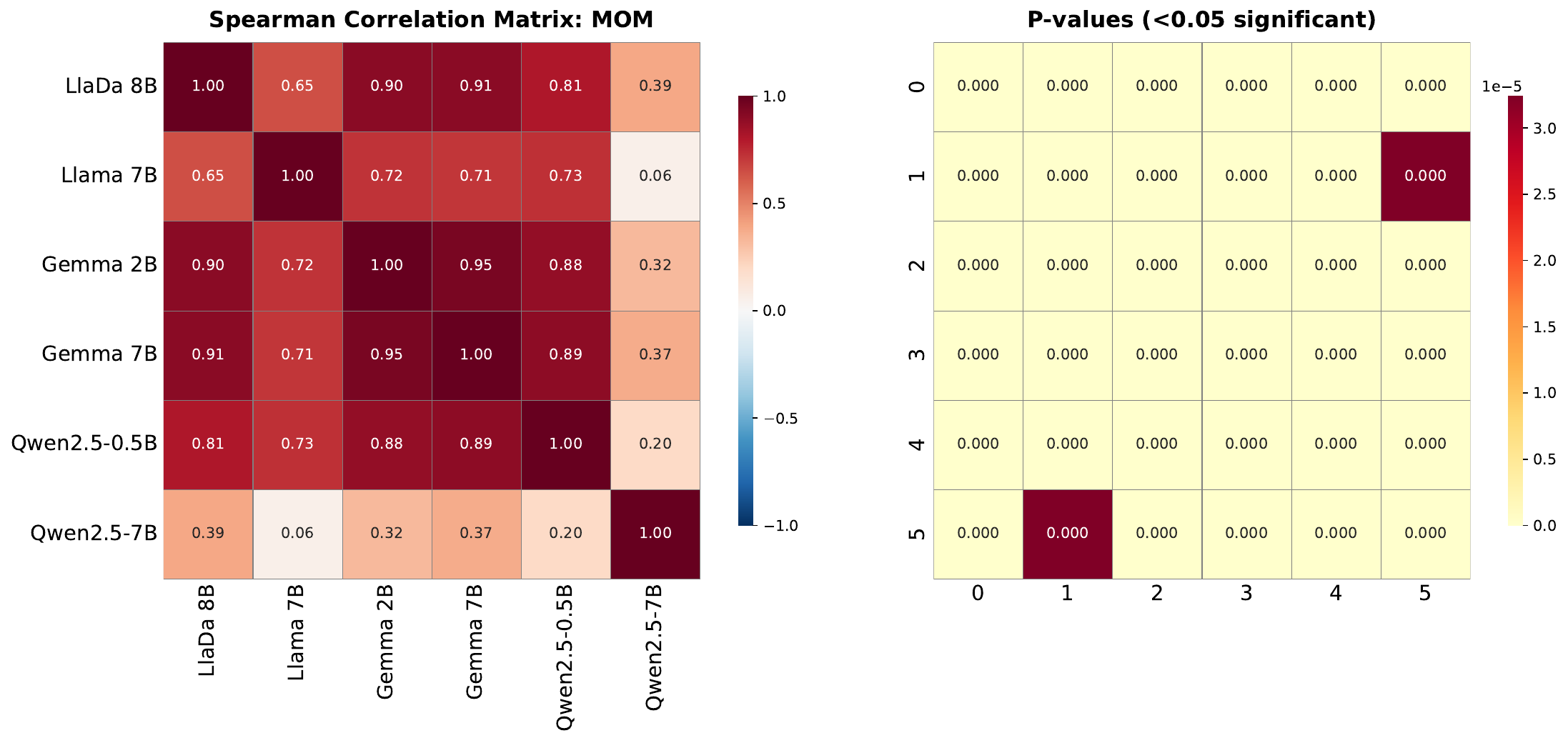}
    \caption{The correlation analysis of quantized models for the following generator models: LLama AWQ (4 bit), LLama (fp32), Mistral AWQ (4 bit), Mistral (fp32), Mistral AQLM (2 bit)}
    \label{fig:cor_quan_MOM}
\end{figure}

\begin{figure}[ht]
\centering
    \includegraphics[width=0.9\linewidth]{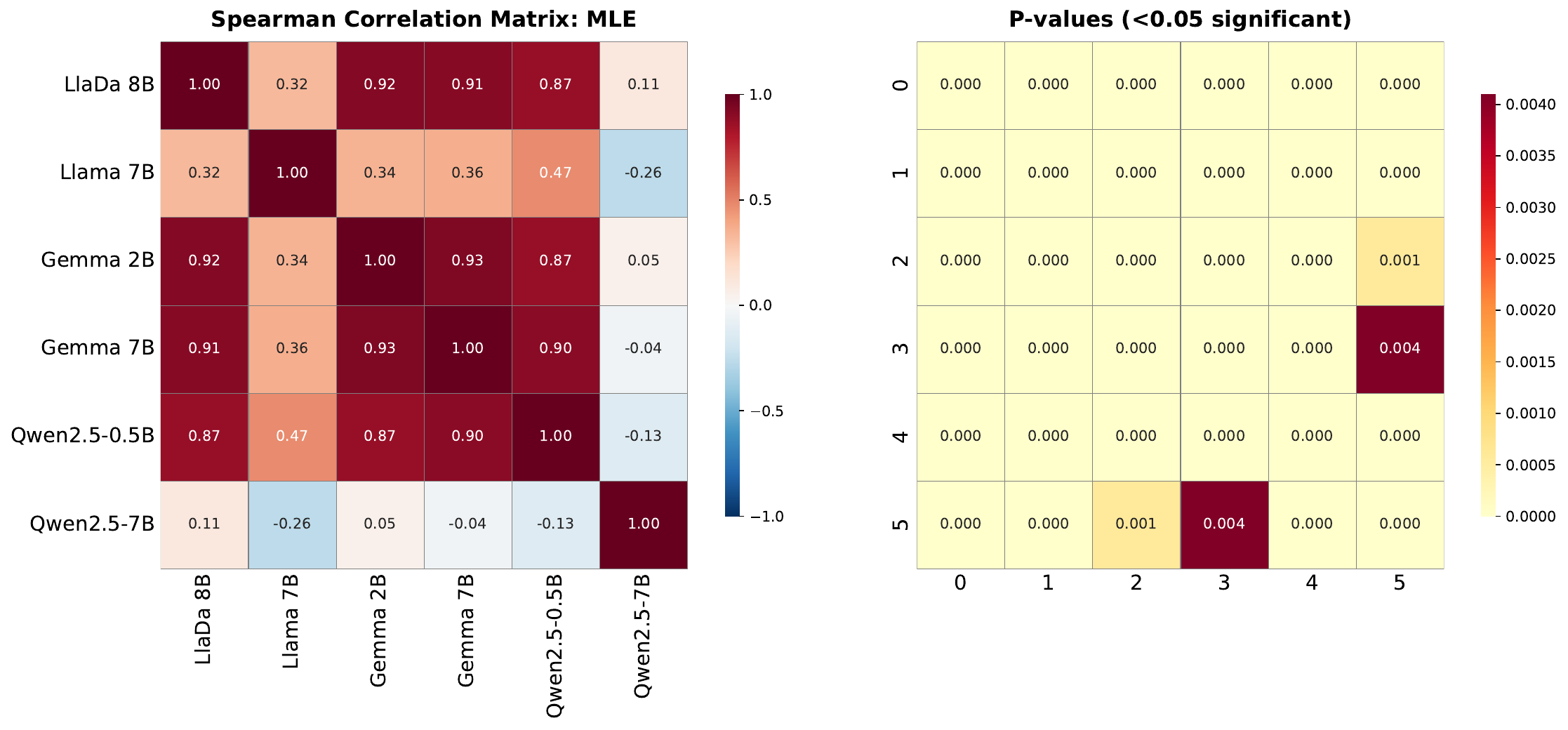}
    \caption{The correlation analysis of quantized models for the following generator models: LLama AWQ (4 bit), LLama (fp32), Mistral AWQ (4 bit), Mistral (fp32), Mistral AQLM (2 bit)}
    \label{fig:cor_quan_MLE}
\end{figure}

In this subsection, we study the average and layer-wise metrics for Quantized models study for LLama and Mistral models for different quantization techniques for $6$ Ntster models $\mathcal{T}$ (see Figures \ref{fig:quantized_total_1} and \ref{fig:quantized_total_2} for average metrics, \ref{fig:quantized_layer_1} and \ref{fig:quantized_layer_2} for layer-wise metrics and correlation metrics \ref{fig:cor_quan_Schatten Norm}, \ref{fig:cor_quan_MOM}, \ref{fig:cor_quan_MLE})

\newpage
\section{Extra Results}
\label{appendix}
In the following appendix sections, we present additional experiments that include various metrics $\mathbf{R}$, tester models $\mathcal{T}$, and generator models $\mathcal{G}$. The metrics we utilize as $\mathbf{R}$ are Schatten Norms, Maximum Explainable Variance, Effective Rank, Resultant Length, MAUVE, and Intrinsic Dimensionality metrics (MLE, MOM, MADA, and CorrInt).

For the Tester models $\mathcal{T}$ we use Gemma-1-7b, Gemma-2b-it, LLama-3.1-8B-it, Qwen-2.5-7b-it, Qwen-2-0.5B and LLaDa-8B.

For the generator models $\mathcal{G}$ we utilize Gemma-2b-it, Qwen-2.5-7b-it, LLama-3.1-8B-it, Deepseek-R1, Mistral-7b-it, Phi-3-medium-4k-it, Phi-3-mini-128k-it and Starling-lm-7b-beta. 

\begin{figure}
    \centering
    \includegraphics[width=1\linewidth]{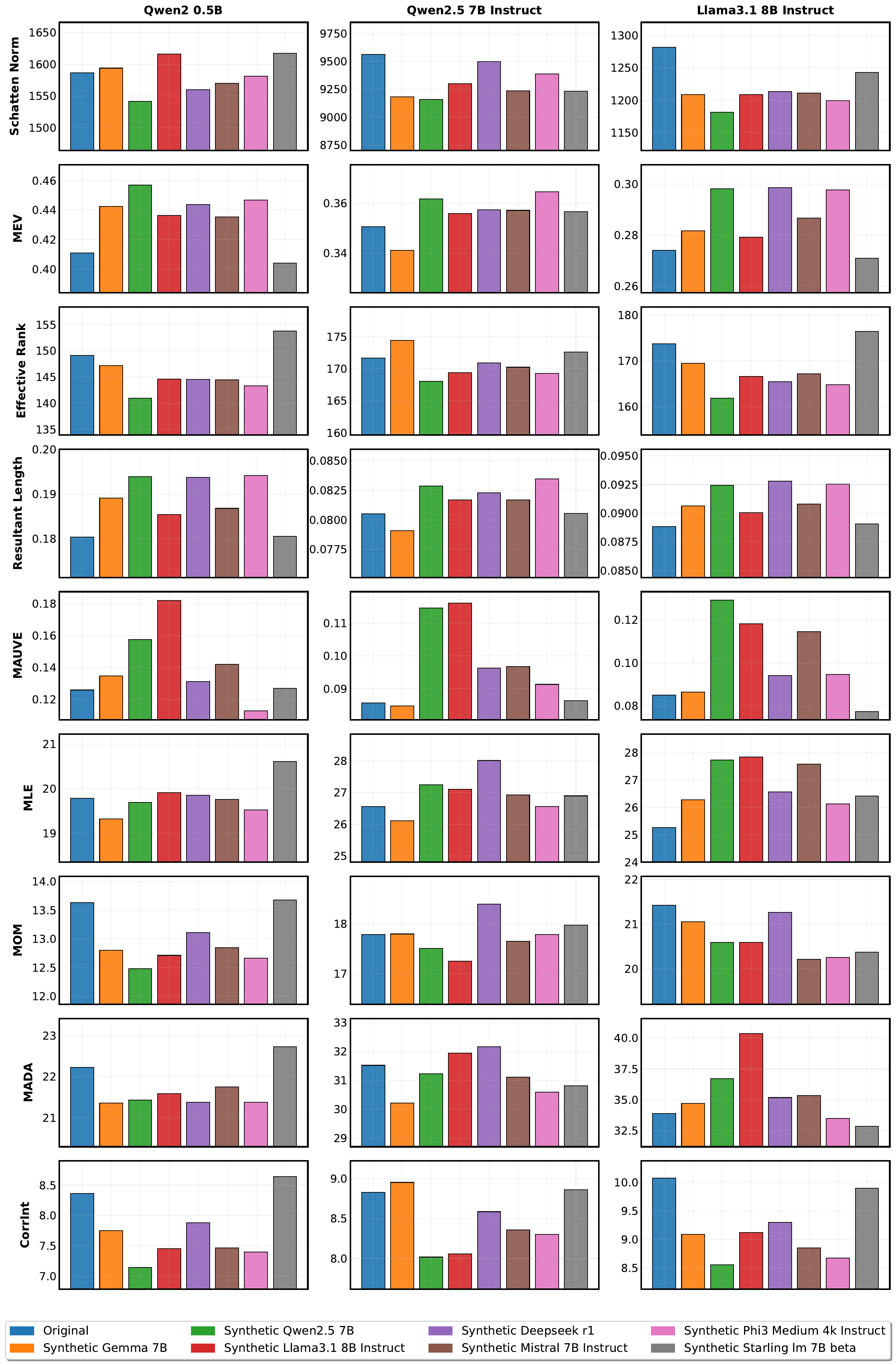}
    \caption{The ranking of eight generators models via tester models (Qwen2 0.5B, Qwen2.5 7B Instruct and LLama3.1 8B Instruct) and all geometric metrics.}
    \label{fig:total_1}
\end{figure}

\begin{figure}
    \centering
    \includegraphics[width=1\linewidth]{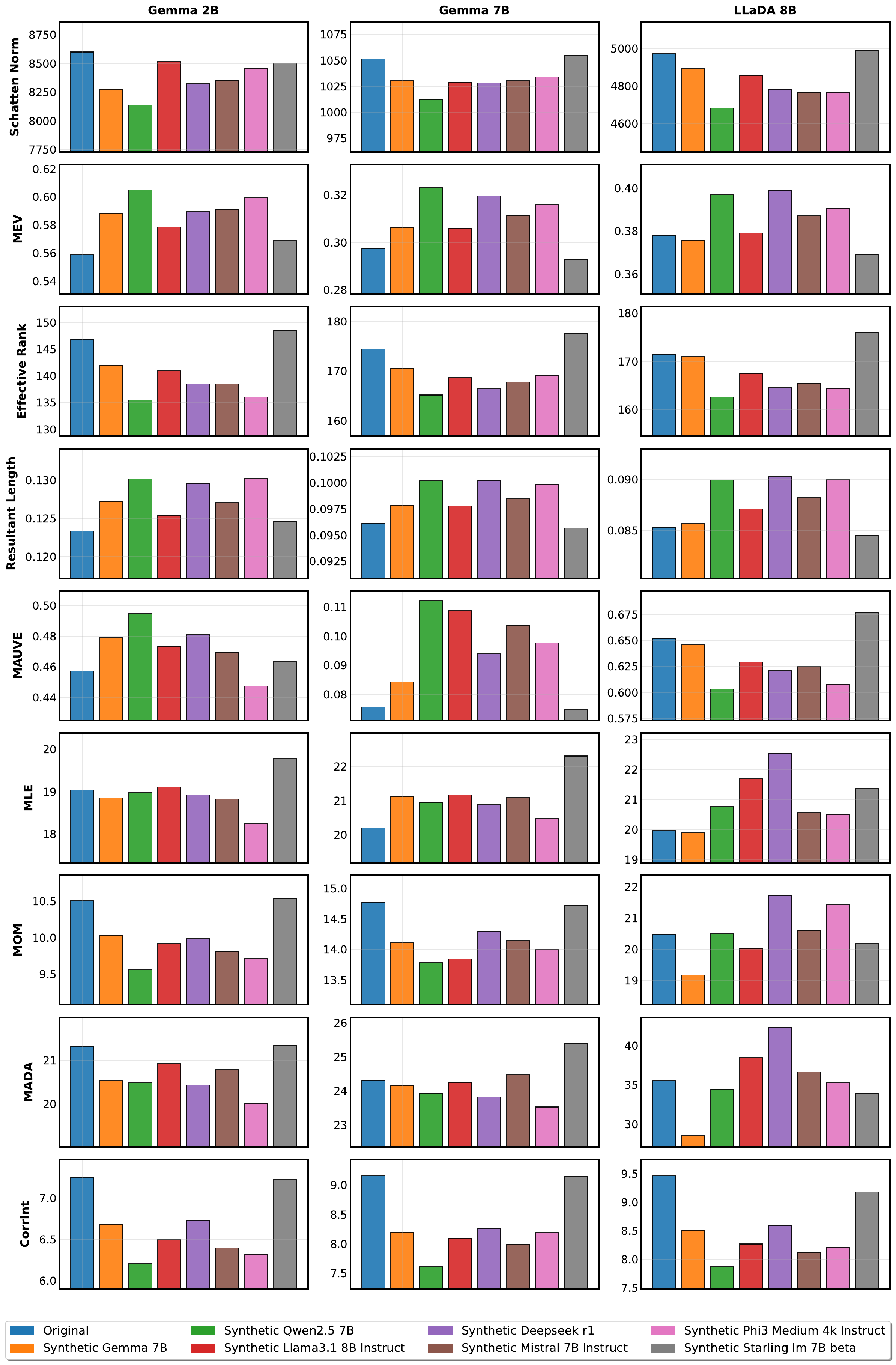}
    \caption{The ranking of eight generators models via tester models (Gemma 2B, Gemma 7B and LLaDA 8B) and all geometric metrics.}
    \label{fig:total_2}
\end{figure}

\begin{figure}
    \centering
    \includegraphics[width=1\linewidth]{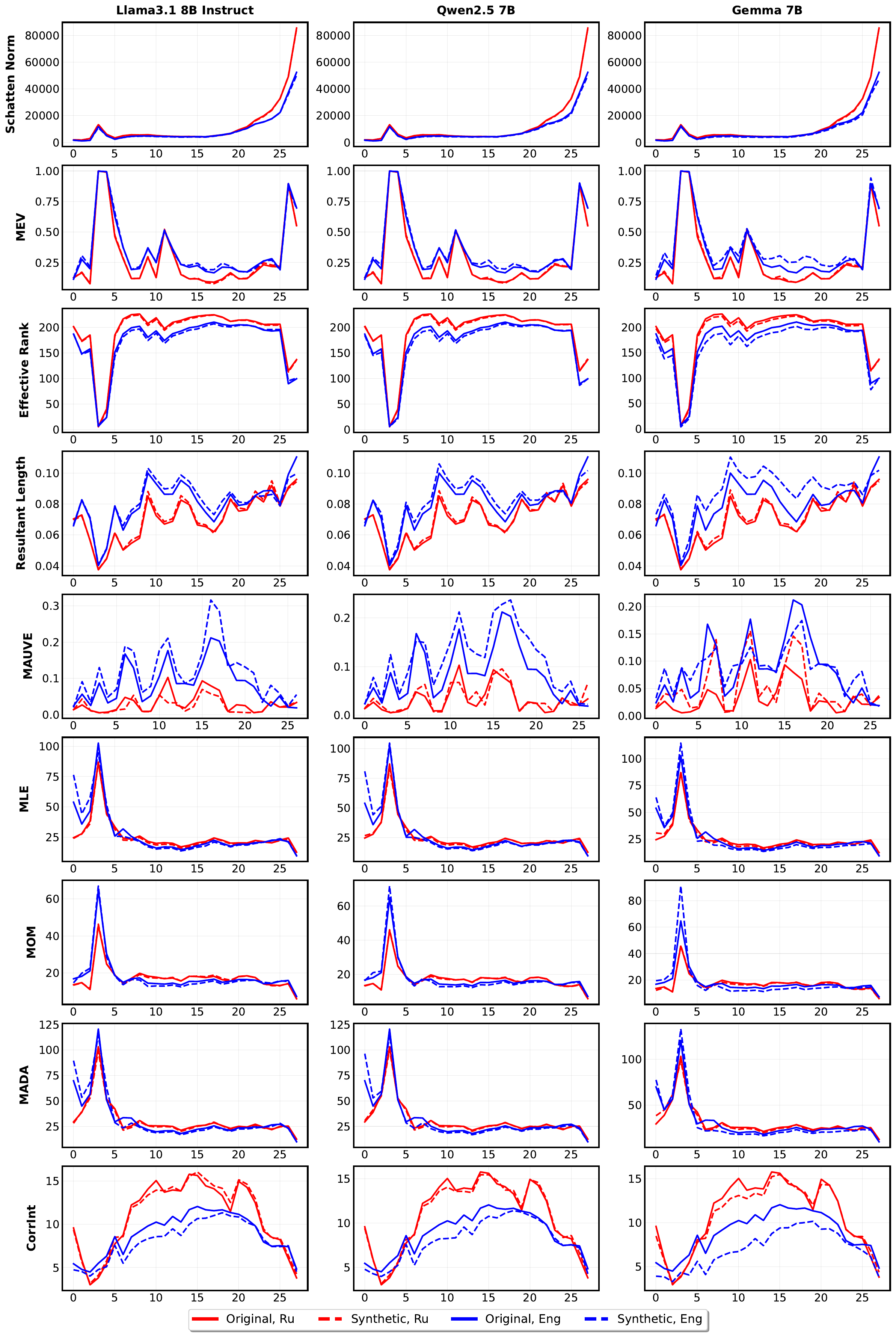}
    \caption{The comparison of Original and Synthetic Russian and English texts generated by Llama3.1 8B Instruct, Qwen2.5 7B Instruct and Gemma 2B and tested via Qwen2.5 7B Instruct.}
    \label{fig:languages_app_2}
\end{figure}

\begin{figure}
    \centering
    \includegraphics[width=1\linewidth]{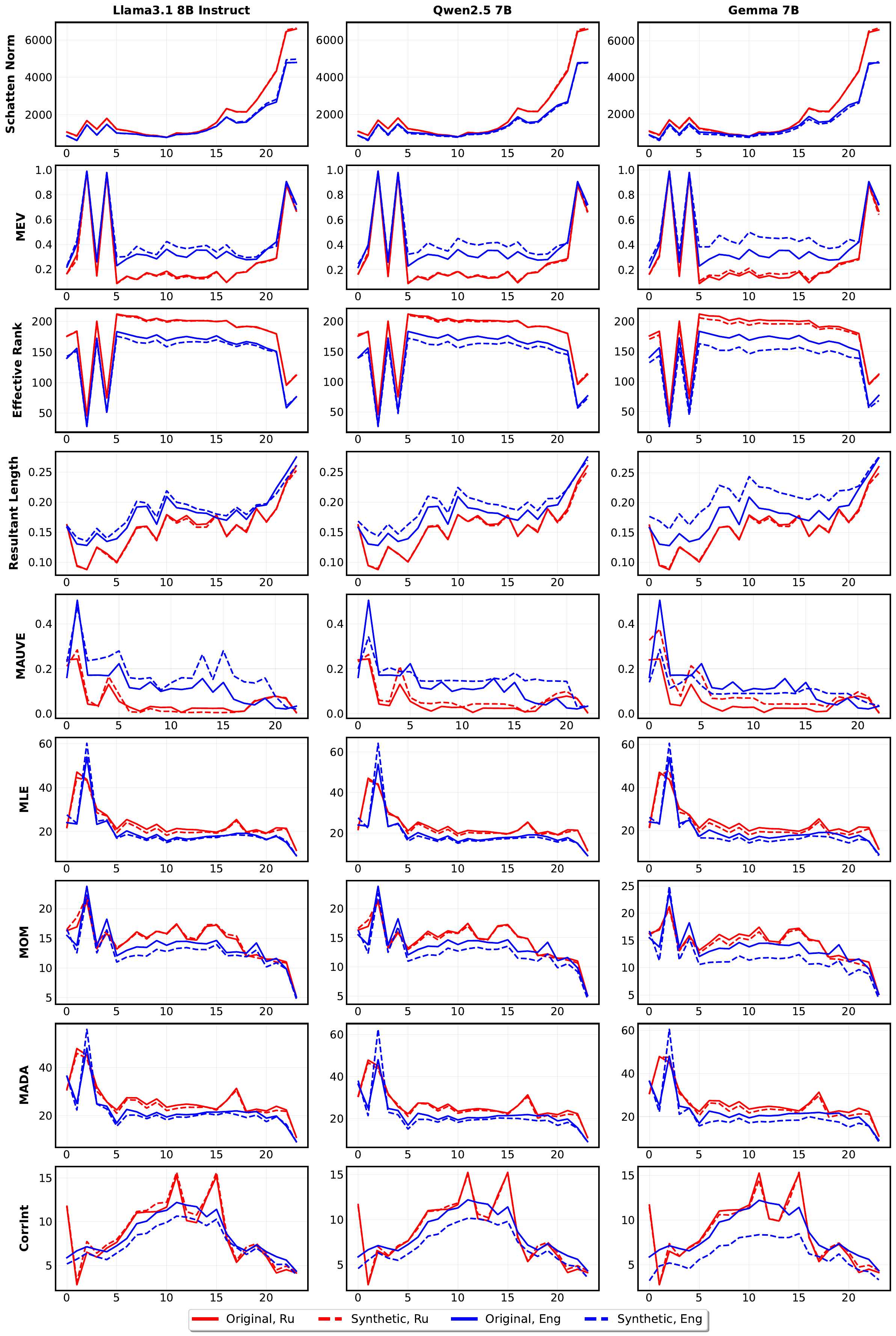}
    \caption{The comparison of Original and Synthetic Russian and English texts generated by Llama3.1 8B Instruct, Qwen2.5 7B Instruct and Gemma 2B and tested via Qwen2 0.5B.}
    \label{fig:languages_app_3}
\end{figure}

\begin{figure}
    \centering
    \includegraphics[width=1\linewidth]{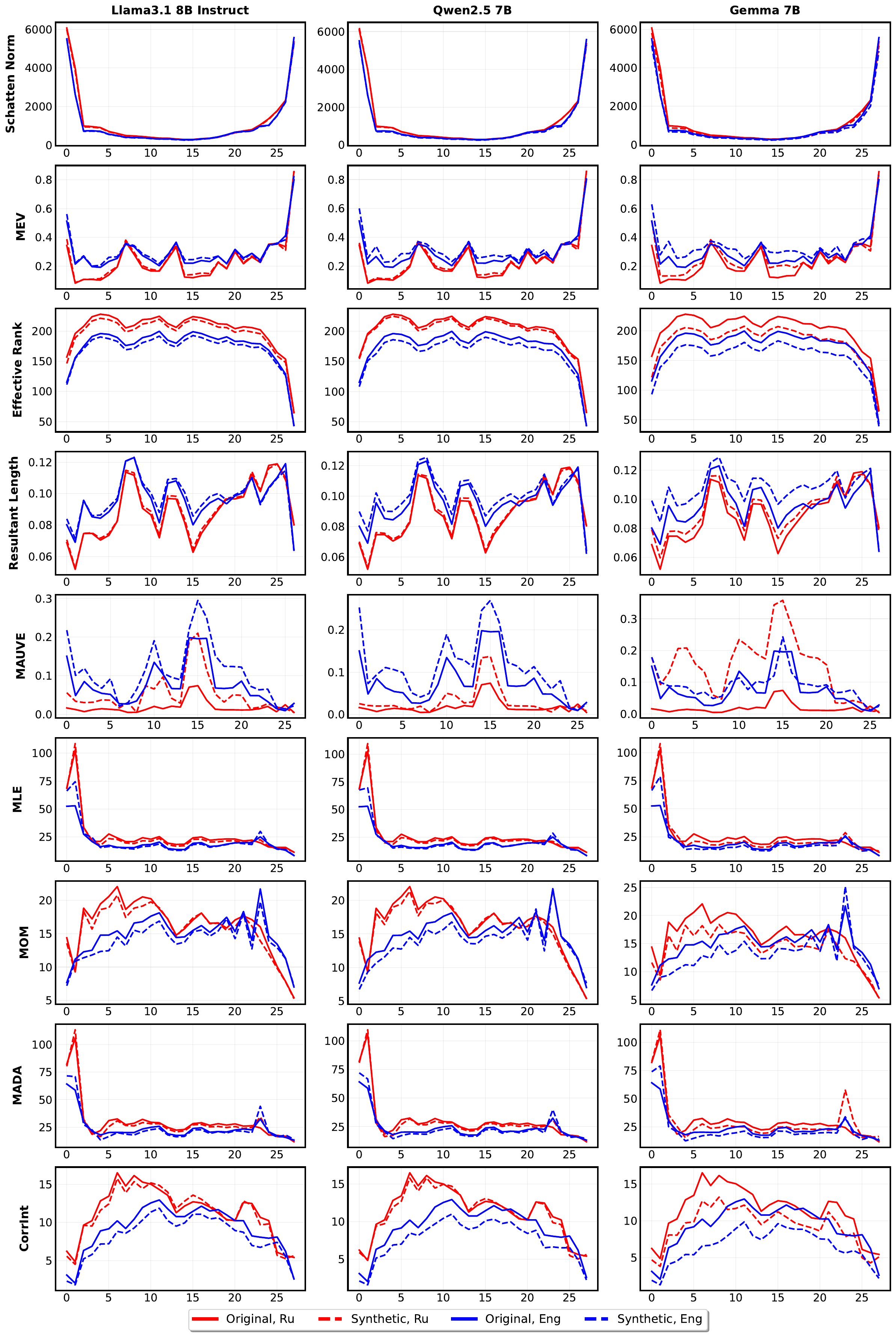}
    \caption{The comparison of Original and Synthetic Russian and English texts generated by Llama3.1 8B Instruct, Qwen2.5 7B Instruct and Gemma 2B and tested via Gemma 7B.}
    \label{fig:languages_app_4}
\end{figure}

\begin{figure}
    \centering
    \includegraphics[width=1\linewidth]{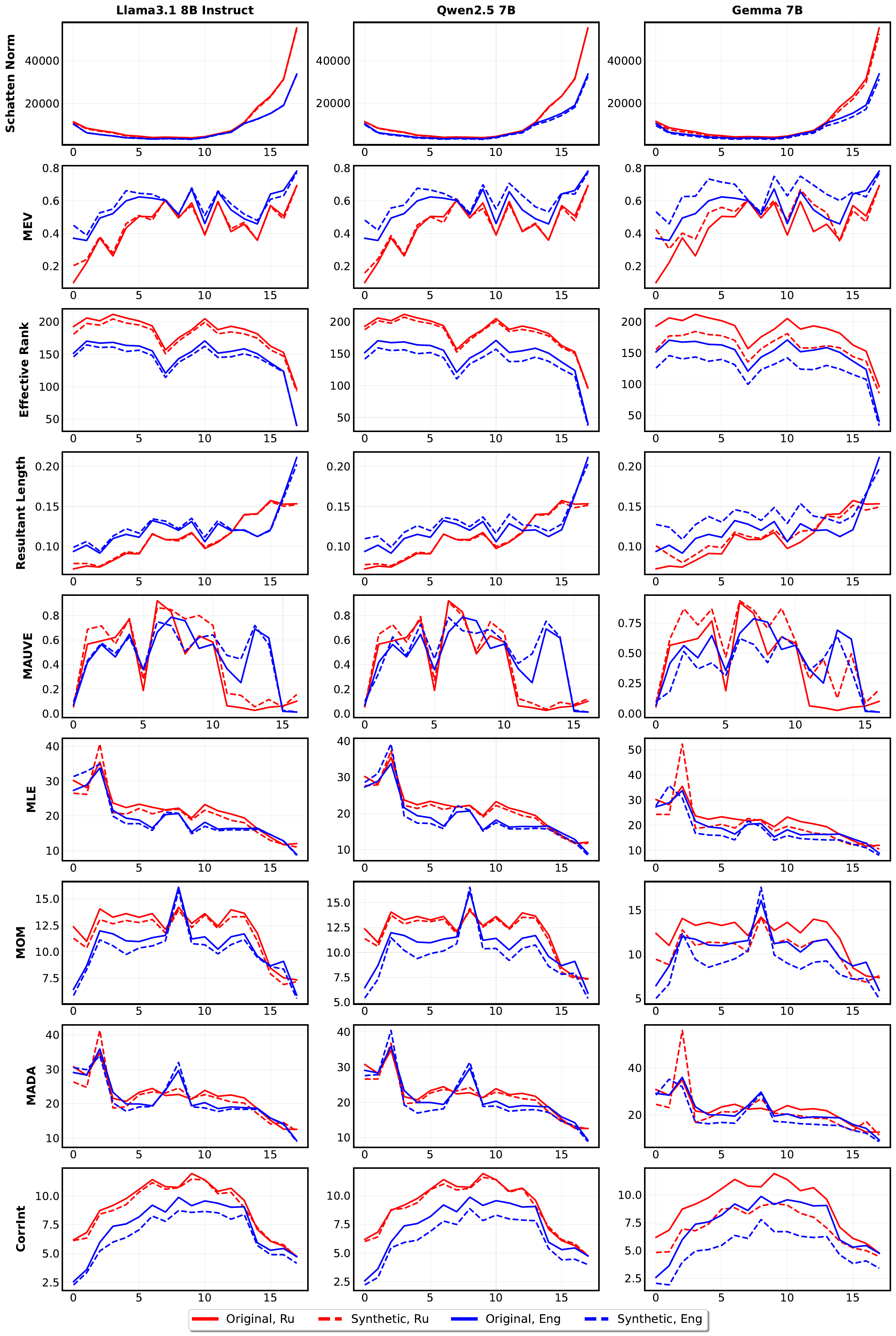}
    \caption{The comparison of Original and Synthetic Russian and English texts generated by Llama3.1 8B Instruct, Qwen2.5 7B Instruct and Gemma 2B and tested via Gemma 2B.}
    \label{fig:languages_app_5}
\end{figure}

\begin{figure}
    \centering
    \includegraphics[width=1\linewidth]{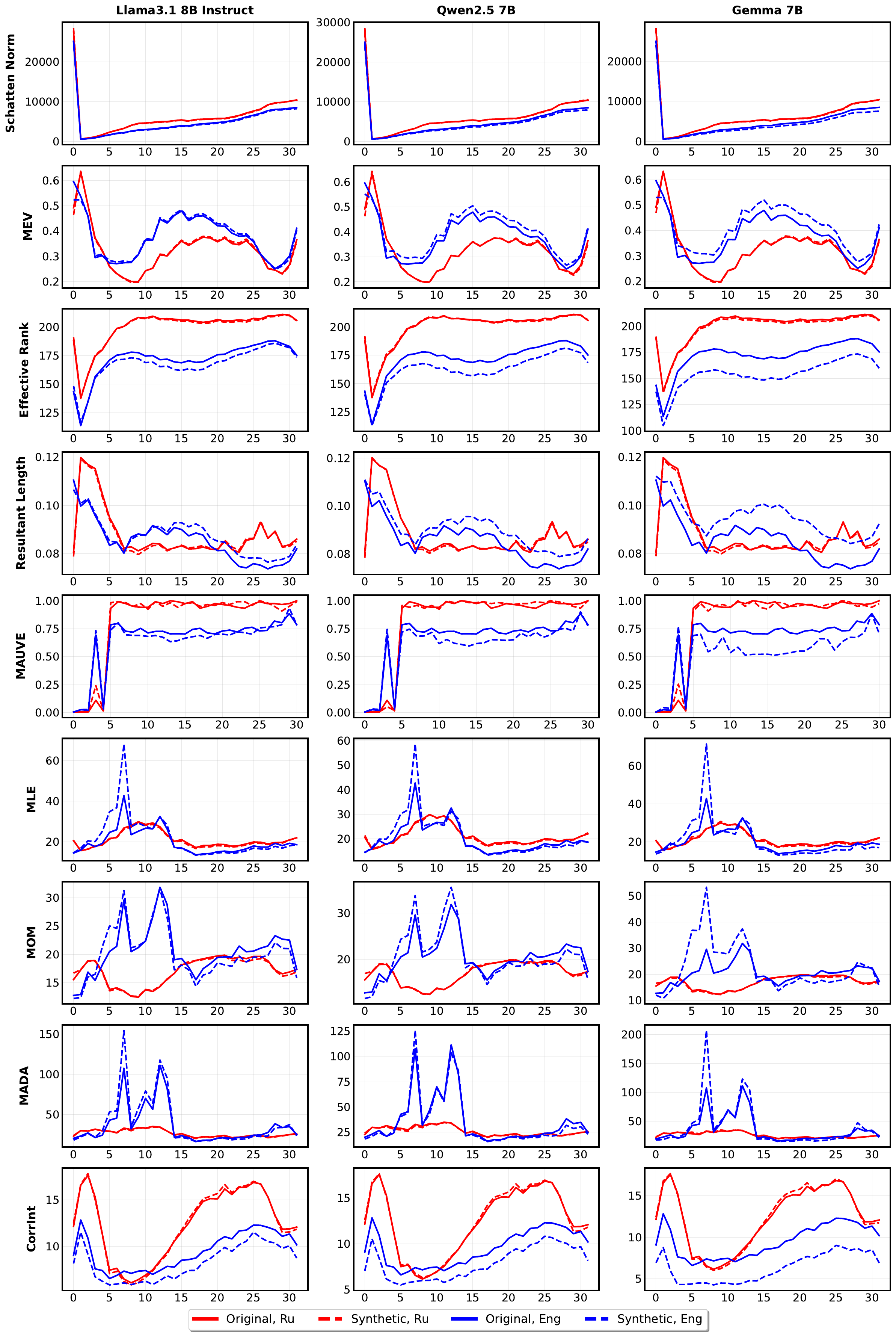}
    \caption{The comparison of Original and Synthetic Russian and English texts generated by Llama3.1 8B Instruct, Qwen2.5 7B Instruct and Gemma 2B and tested via LLaDA 8B.}
    \label{fig:languages_app_6}
\end{figure}

\begin{figure}
    \centering
    \includegraphics[width=1\linewidth]{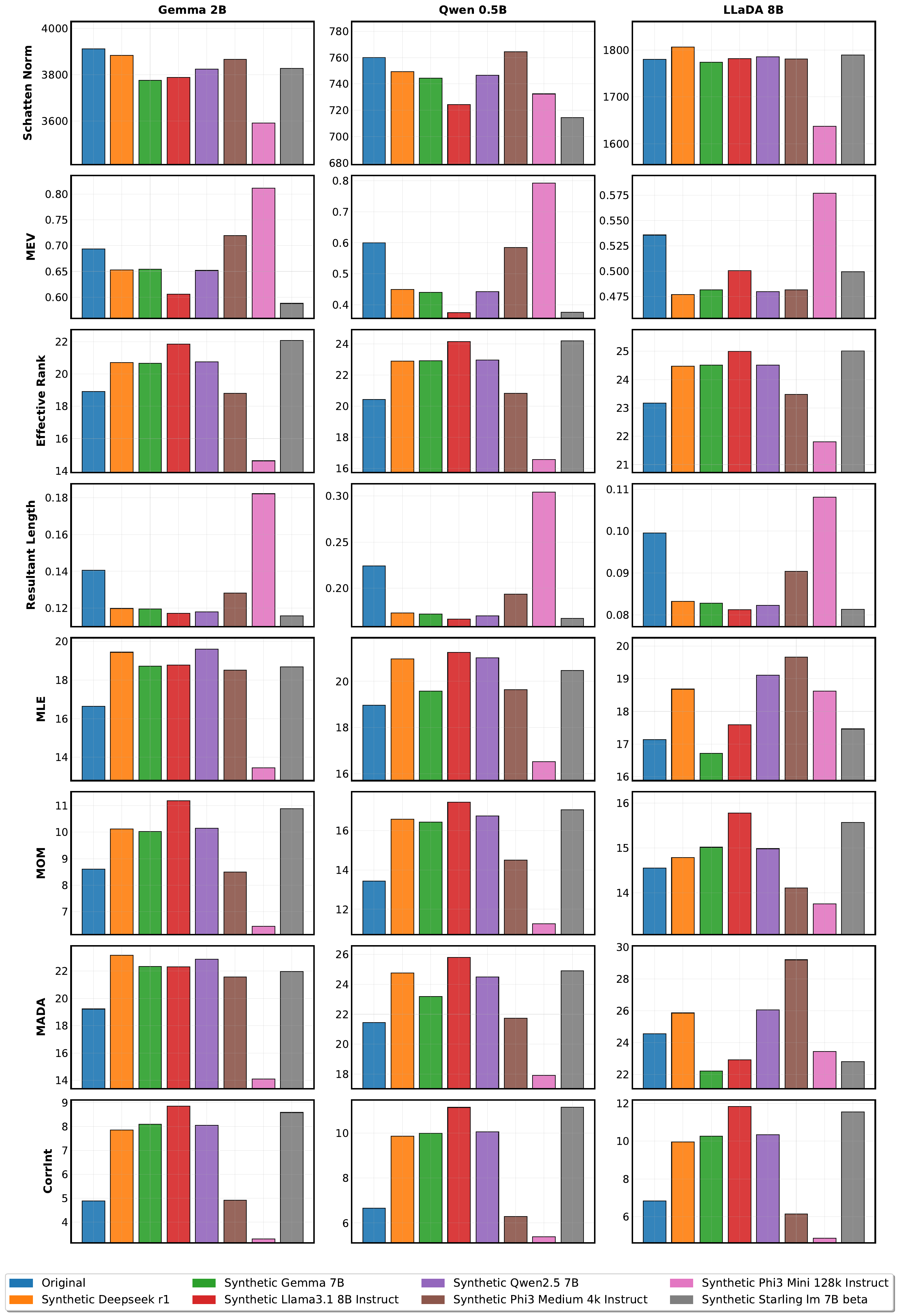}
    \caption{The ranking of seven generators models via tester models (Qwen2 0.5B, Gemma 2B and LLaDA 8B) and all geometric metrics. The models rewrite the text.}
    \label{fig:gen_avg_1}
\end{figure}

\begin{figure}
    \centering
    \includegraphics[width=1\linewidth]{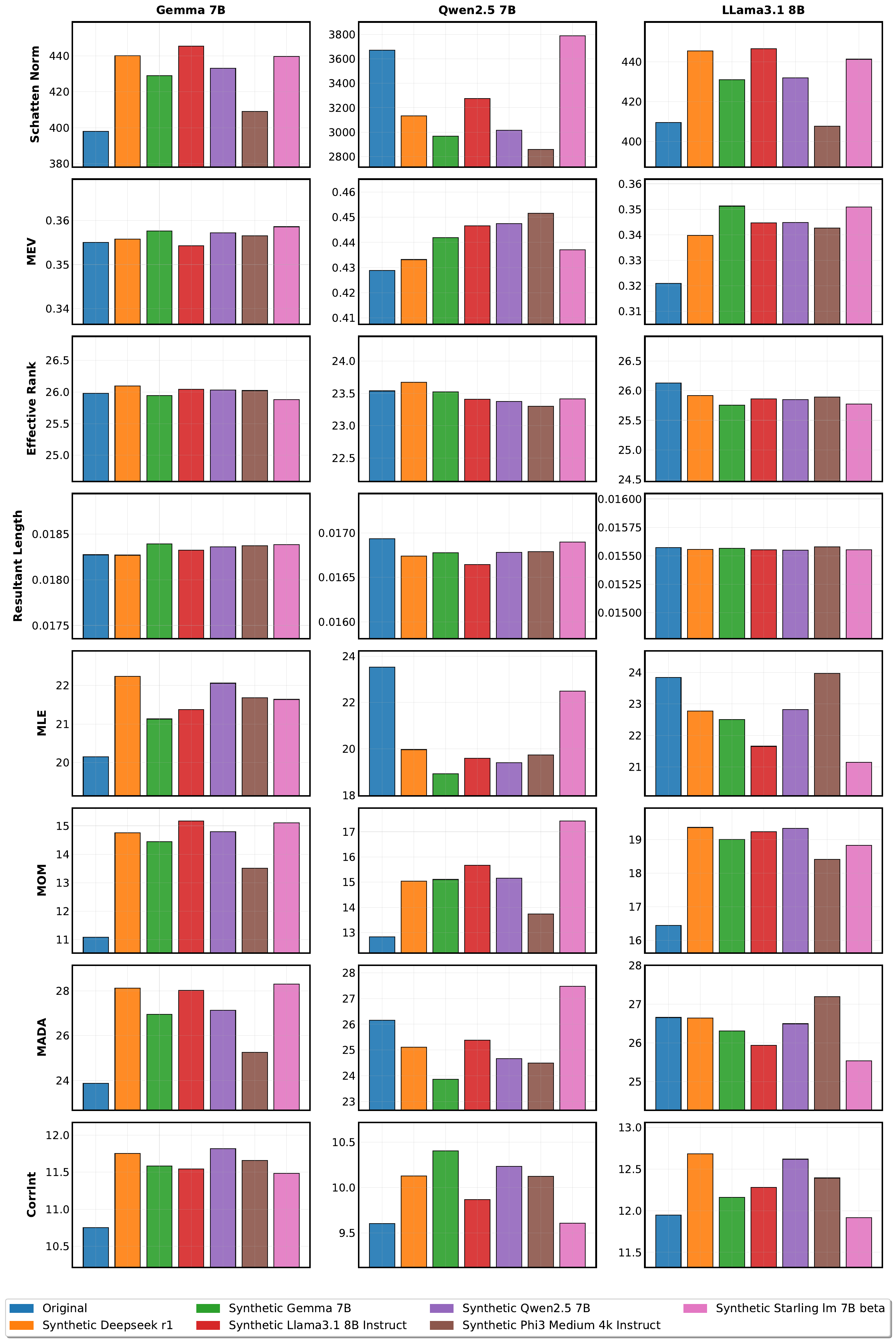}
    \caption{The  ranking of seven generators models via tester models (Qwen2.5 7B Instruct, Gemma 7B and Llama3.1 8B Instruct) and all geometric metrics. The models rewrite the text.}
    \label{fig:gen_avg_2}
\end{figure}

\begin{figure}
    \centering
    \includegraphics[width=1\linewidth]{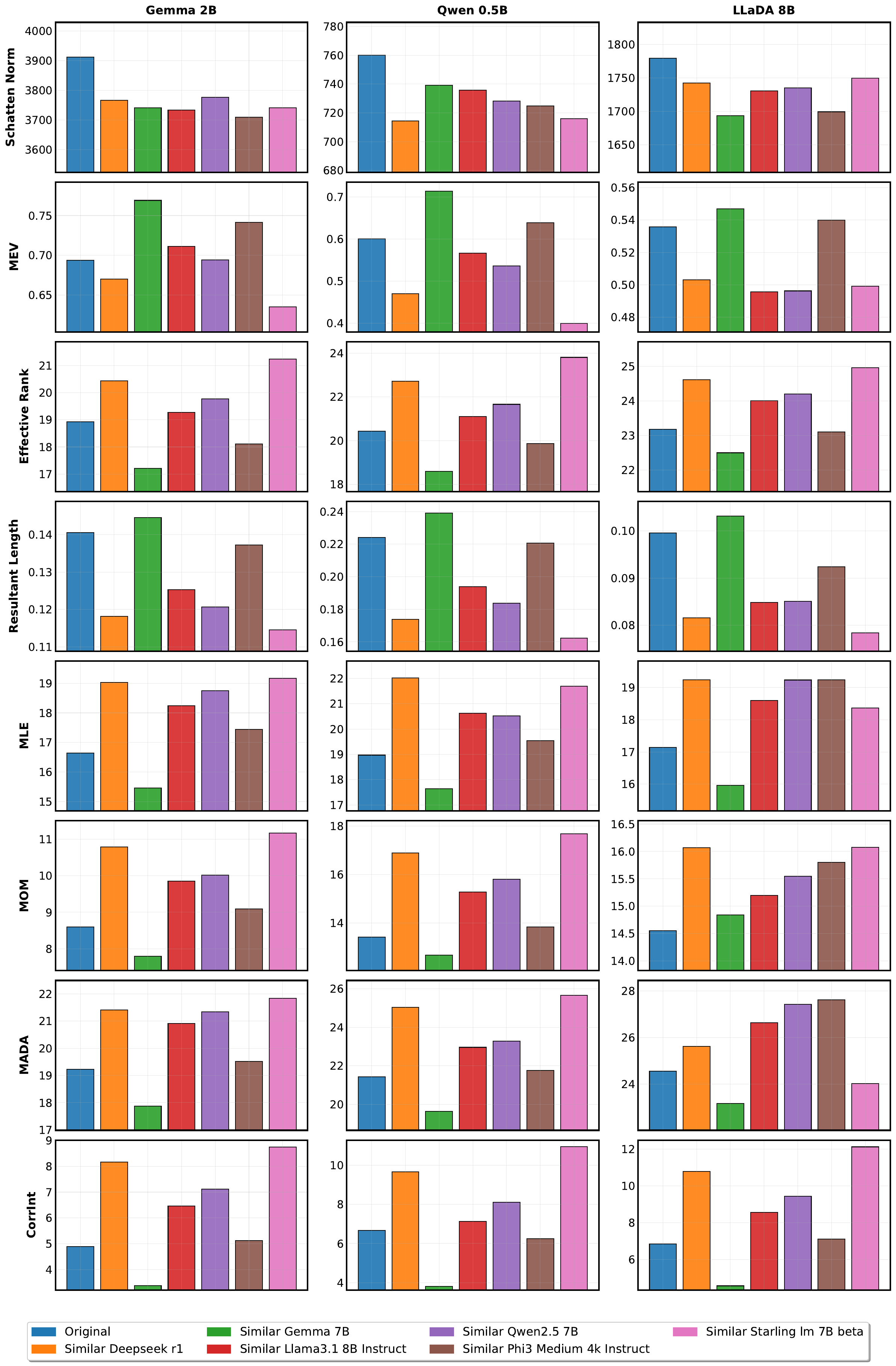}
    \caption{The  ranking of seven generators models via tester models (Qwen2 0.5B, Gemma 2B and LLaDA 8B) and all geometric metrics. The models simplify the text.}
    \label{fig:sim_avg_1}
\end{figure}

\begin{figure}
    \centering
    \includegraphics[width=1\linewidth]{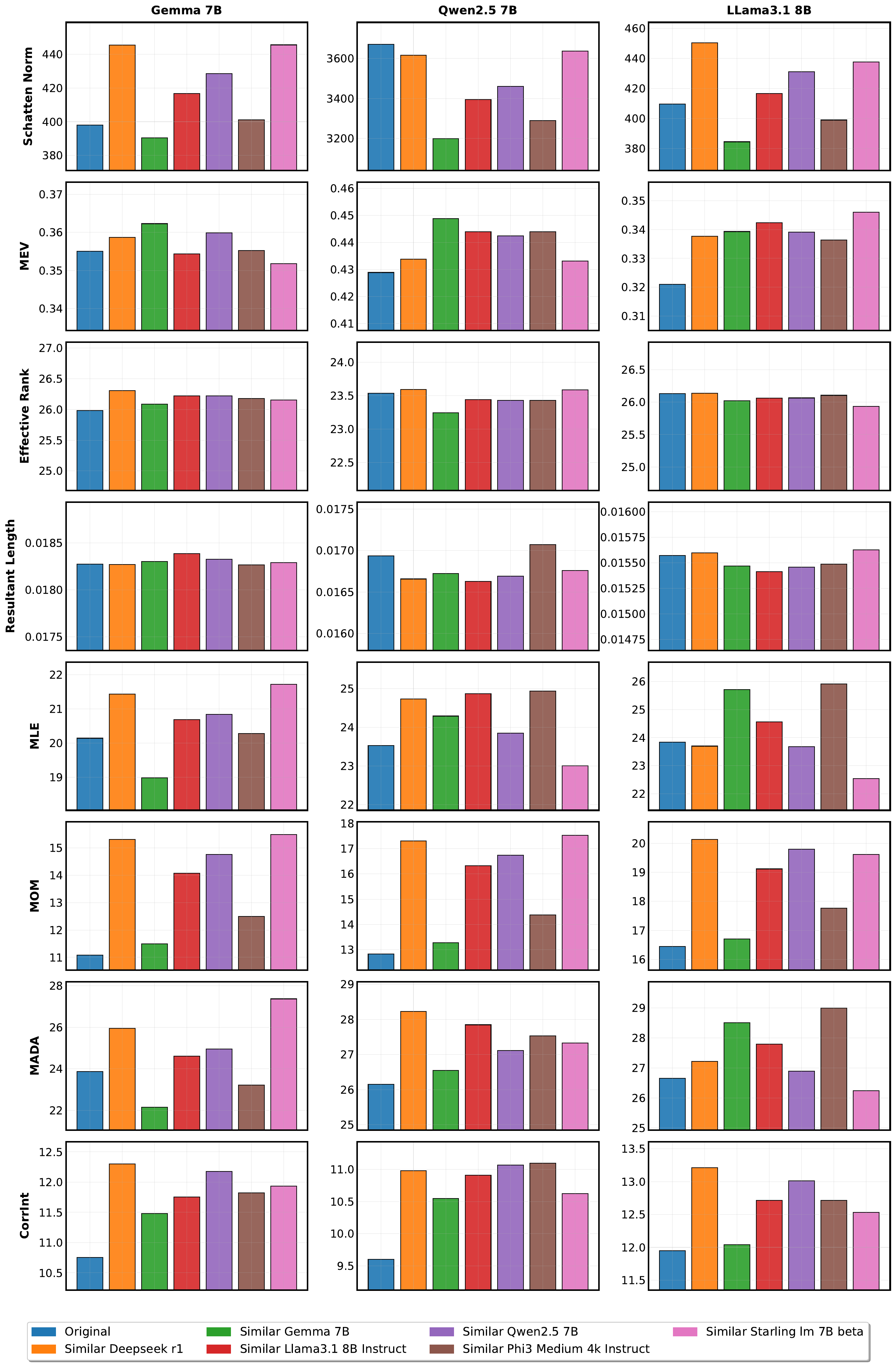}
    \caption{The  ranking of seven generators models via tester models (Qwen2.5 7B Instruct, Gemma 7B and Llama3.1 8B Instruct) and all geometric metrics. The models simplify the text.}
    \label{fig:sim_avg_2}
\end{figure}

\begin{figure}
    \centering
    \includegraphics[width=1\linewidth]{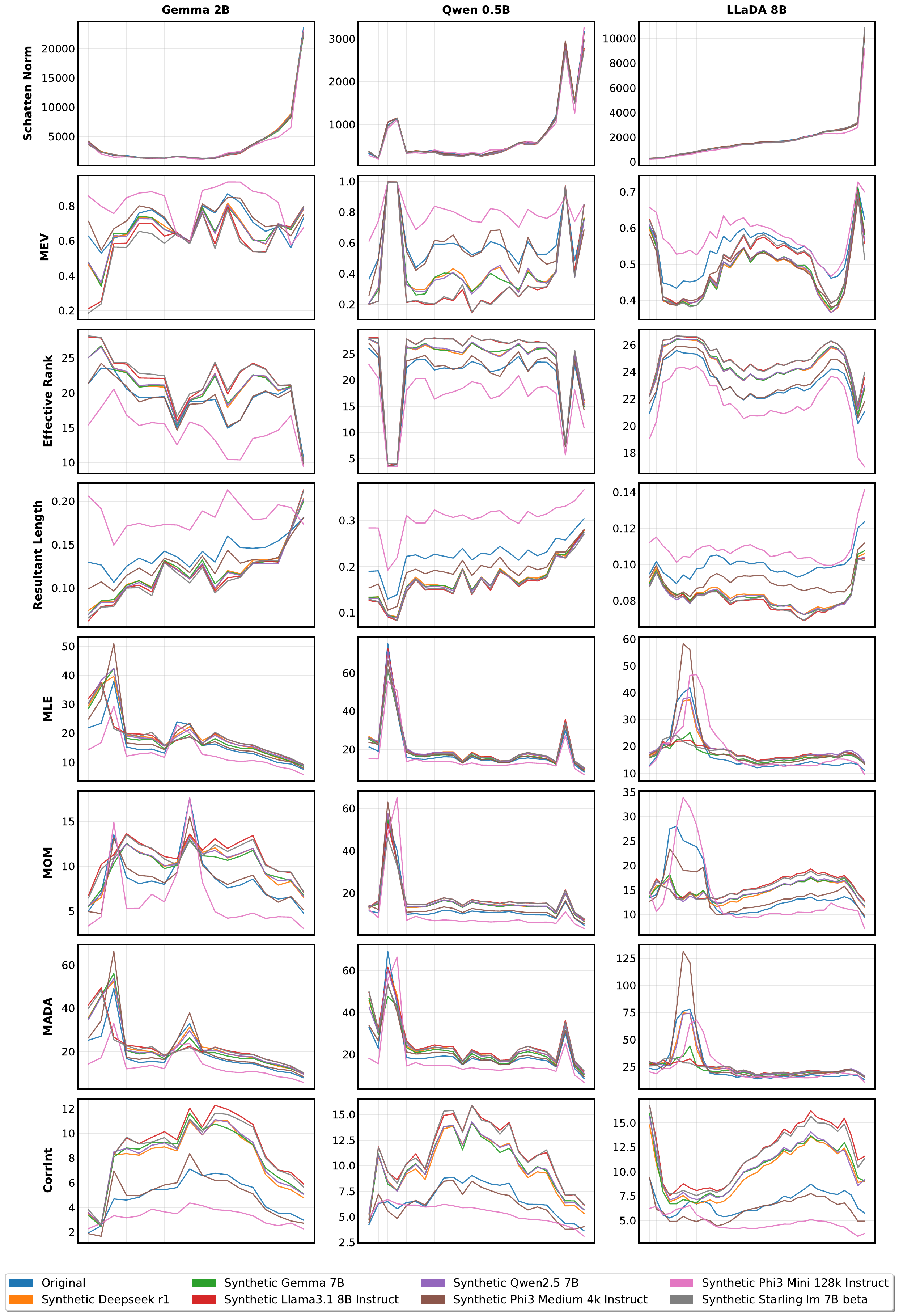}
    \caption{The layer-wise ranking of seven generators models via tester models (Qwen2 0.5B, Gemma 2B and LLaDA 8B) and all geometric metrics. The models rewrite the text.}
    \label{fig:gen_layer_1}
\end{figure}

\begin{figure}
    \centering
    \includegraphics[width=1\linewidth]{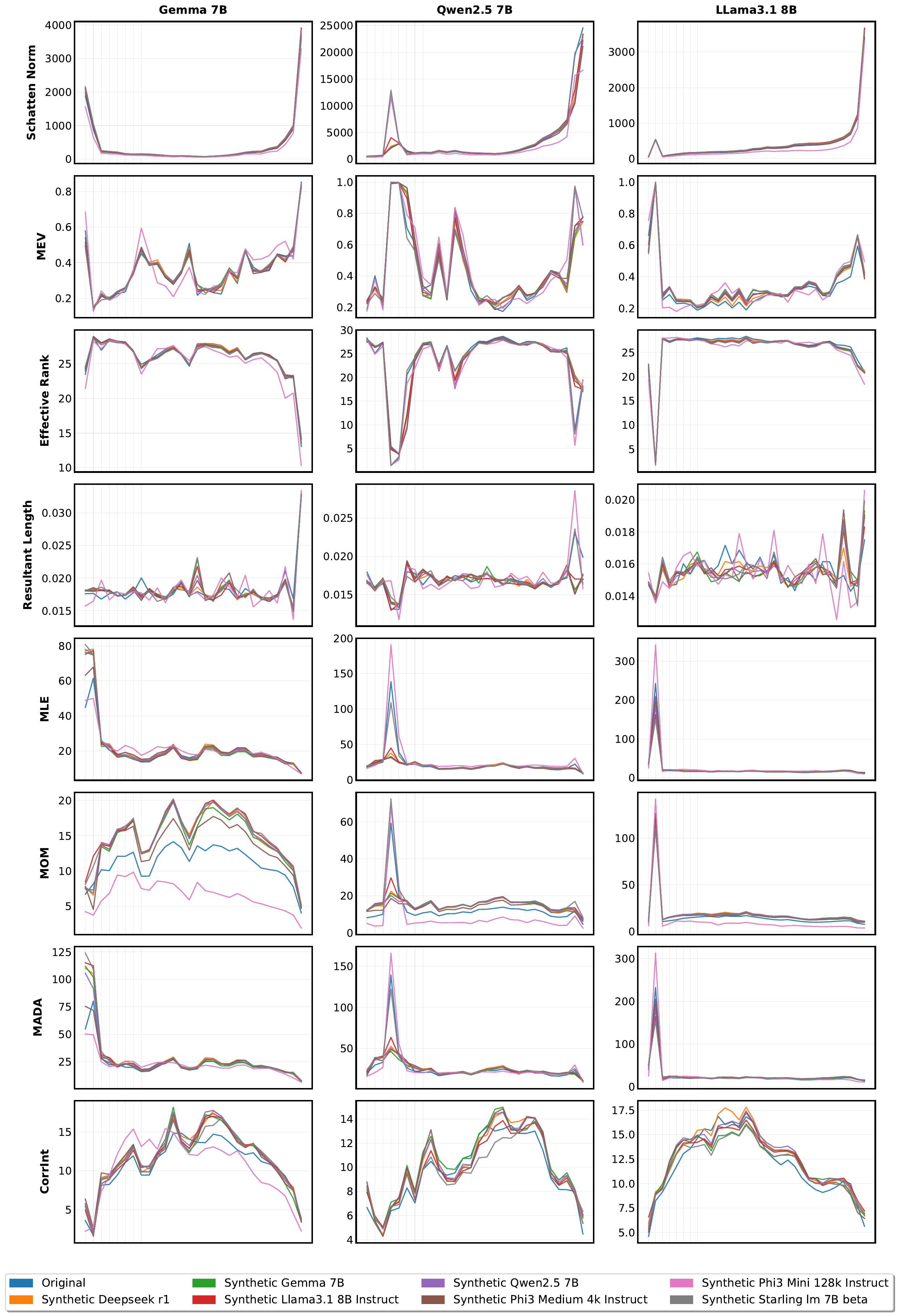}
    \caption{The layer-wise ranking of seven generators models via tester models (Qwen2.5 7B Instruct, Gemma 7B and Llama3.1 8B Instruct) and all geometric metrics. The models rewrite the text.}
    \label{fig:gen_layer_2}
\end{figure}

\begin{figure}
    \centering
    \includegraphics[width=1\linewidth]{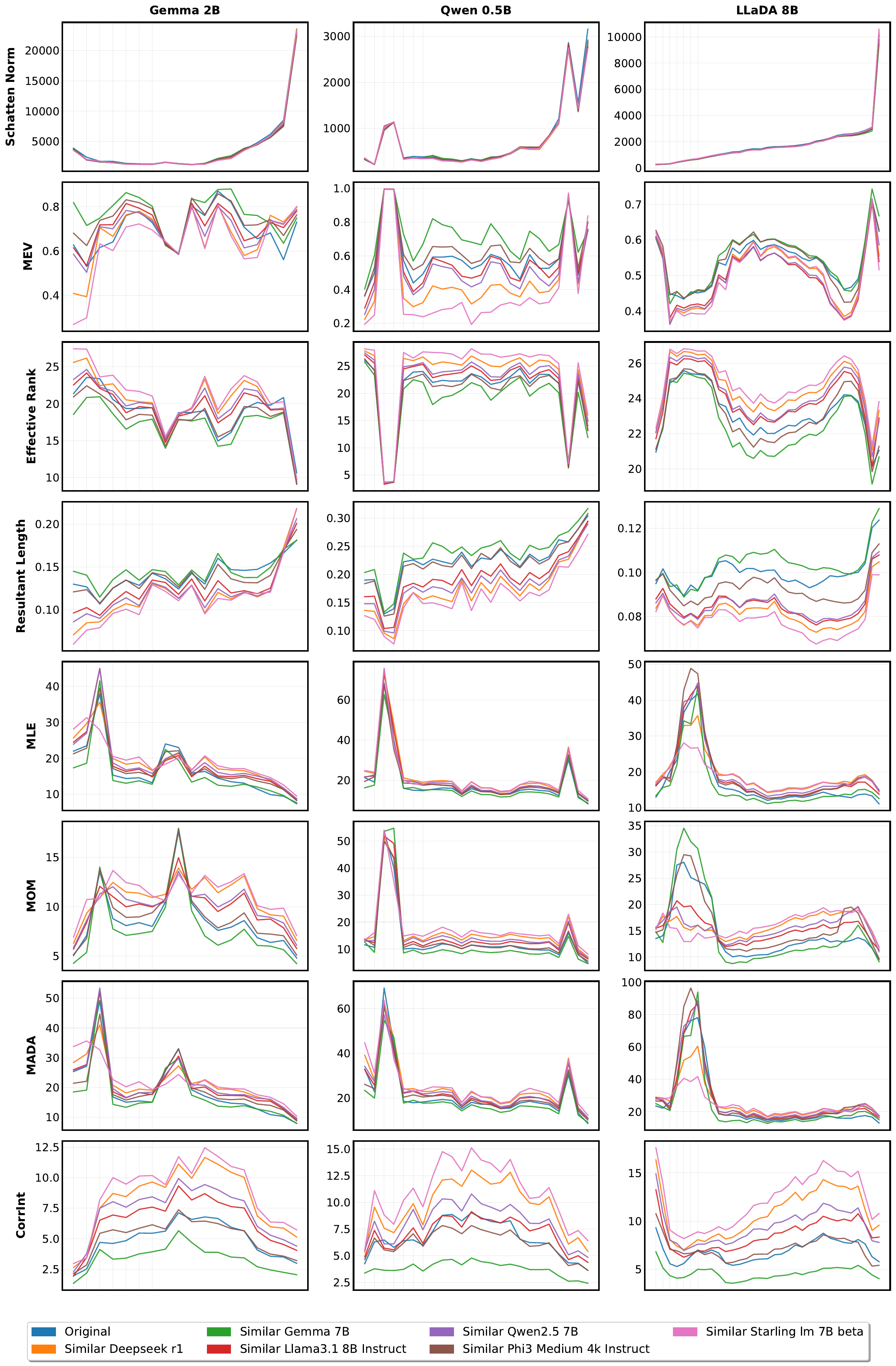}
    \caption{The layer-wise ranking of seven generators models via tester models (Qwen2 0.5B, Gemma 2B and LLaDA 8B) and all geometric metrics. The models simplify the text.}
    \label{fig:sim_layer_1}
\end{figure}

\begin{figure}
    \centering
    \includegraphics[width=1\linewidth]{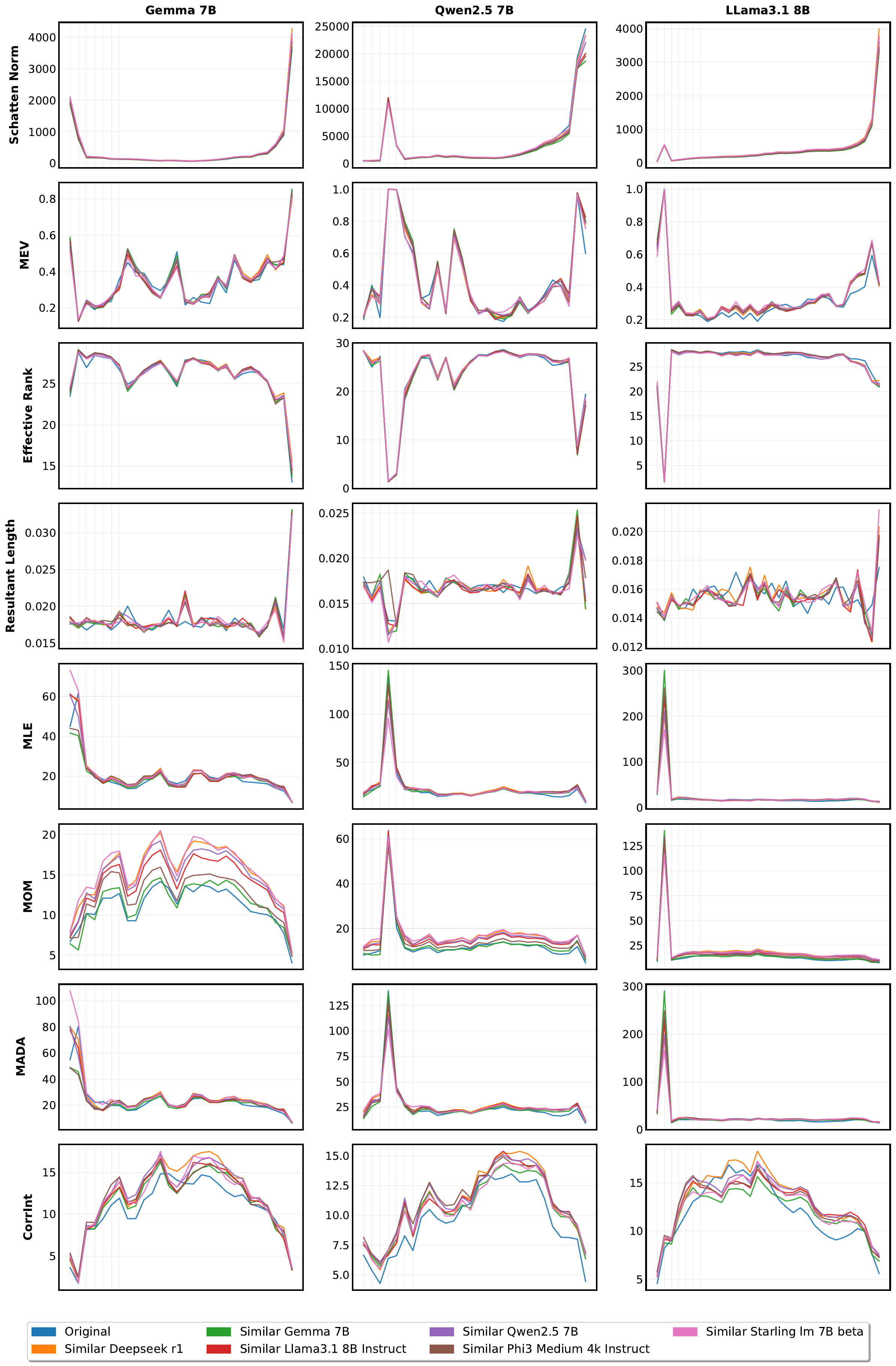}
    \caption{The layer-wise ranking of seven generators models via tester models (Qwen2.5 7B Instruct, Gemma 7B and Llama3.1 8B Instruct) and all geometric metrics. The models simplify the text.}
    \label{fig:sim_layer_2}
\end{figure}

The structure of the section is:
\begin{itemize}
    \item The average metrics for Movie Reviews for different tester $\mathcal{T}$ and generator $\mathcal{G}$ models are demonstrated in Figures \ref{fig:total_1} and \ref{fig:total_2}.
    
    \item For the metrics for Movie Reviews for different layers of tester $\mathcal{T}$ models for various generator $\mathcal{G}$ models see Figures \ref{fig:layer_profiles_1} and \ref{fig:layer_profiles_2} in the main text.
    
    \item The layer-wise metrics for Russian and English languages and different tester and generator models are demonstrated in Figure~\ref{fig:crosslingual_layers} (main text) and Figures \ref{fig:languages_app_2}, \ref{fig:languages_app_3}, \ref{fig:languages_app_4}, \ref{fig:languages_app_5} and \ref{fig:languages_app_6}.
    
    \item The average metrics for Generated from title abstracts for different tester $\mathcal{T}$ and generator $\mathcal{G}$ models are demonstrated in Figures \ref{fig:gen_avg_1} and \ref{fig:gen_avg_2}.
    
    \item For the metrics for Generated from title abstracts for different layers of tester $\mathcal{T}$ models for various generator $\mathcal{G}$ models see Figures \ref{fig:gen_layer_1} and \ref{fig:gen_layer_2}.

    \item The average metrics for Simplified abstracts for different tester $\mathcal{T}$ and generator $\mathcal{G}$ models are demonstrated in Figures \ref{fig:sim_avg_1} and \ref{fig:sim_avg_2}.
    
    \item For the metrics for Simplified abstracts for different layers of tester $\mathcal{T}$ models for various generator $\mathcal{G}$ models see Figures \ref{fig:sim_layer_1} and \ref{fig:sim_layer_2}.

\end{itemize}

Additional plots for Tester models $\mathcal{T}$ correlation and and task comparison could be found in additional ZIP file.


\end{document}